\documentclass[letterpaper]{article} 
\usepackage{aaai23}  
\usepackage{times}  
\usepackage{helvet}  
\usepackage{courier}  
\usepackage[hyphens]{url}  
\usepackage{graphicx} 
\usepackage{natbib}  
\usepackage{caption} 
\usepackage{algorithm}
\usepackage{algorithmic}

\usepackage{newfloat}
\usepackage{listings}
\DeclareCaptionStyle{ruled}{labelfont=normalfont,labelsep=colon,strut=off} 
\floatstyle{ruled}
\newfloat{listing}{tb}{lst}{}
\floatname{listing}{Listing}
\usepackage{comment}
\usepackage{subfigure}
\usepackage{amsthm,amssymb}
\usepackage{textcomp,booktabs}
\usepackage{footmisc}
\usepackage{multirow}
\usepackage{bigstrut}
\usepackage{array}
\usepackage{amsfonts}
\usepackage{bm}
\usepackage[switch]{lineno}
\usepackage{amsmath}
\usepackage{makecell}
\usepackage[table,x11names]{xcolor}

\usepackage[T1]{fontenc}
\usepackage{aecompl}

\newcommand{\sourceimage}{I_s}
\newcommand{\targetimage}{I_t}
\newcommand{\boostimage}{I_b}
\newcommand{\mixtargetimage}{I_{t\oplus s}}
\newcommand{\mixboostimage}{I_{b\oplus s}}

\newcommand{\sourcelabel}{Y_s}
\newcommand{\targetpseudo}{\hat{Y}_b}
\newcommand{\mixlabel}{\hat{Y}_{b\oplus s}}

\newcommand{\sourcelogit}{Z_s}
\newcommand{\mixtargetlogit}{Z_{t\oplus s}}
\newcommand{\mixboostlogit}{Z_{b\oplus s}}
\newcommand{\sourcelogitnorm}{Z^*_s}
\newcommand{\mixtargetlogitnorm}{Z^*_{t\oplus s}}
\newcommand{\mixboostlogitnorm}{Z^*_{b\oplus s}}

\newcommand{\sourceprob}{P^*_s}

\newcommand{\mixtargetprob}{P^*_{t\oplus s}}
\newcommand{\mixboostprob}{P^*_{b\oplus s}}

\newcommand{\loss}{\mathcal{L}_{lc}}
\newcommand{\lossce}{\mathcal{L}_{ce}}
\newcommand{\losssource}{\loss^s}
\newcommand{\losstarget}{\loss^{t\oplus s}}
\newcommand{\lossboost}{\loss^{b\oplus s}}
\newcommand{\losscesource}{\lossce^s}
\newcommand{\losscetarget}{\lossce^{t\oplus s}}
\newcommand{\lossceboost}{\lossce^{b\oplus s}}

\newcommand{\method}{\texttt{VBLC}~}
\newcommand{\methodblank}{\texttt{VBLC}}
\newcommand{\boostModuleNameTitle}{Visibility Boost Module (VBM)}
\newcommand{\lossModuleNameTitle}{Logit-Constraint Learning (LCL)}
\newcommand{\boostModuleName}{\textit{visibility boost module}}
\newcommand{\lossModuleName}{\textit{logit-constraint learning}}

\newcommand{\gr}{\rowcolor[gray]{.85}}
\newcommand{\gc}{\cellcolor[gray]{.85}}

\title{VBLC: Visibility Boosting and Logit-Constraint Learning for \\Domain Adaptive Semantic Segmentation under Adverse Conditions}
\author{
    Mingjia Li\textsuperscript{\rm 1\equalcontrib}, Binhui Xie\textsuperscript{\rm 1\equalcontrib}, Shuang Li\textsuperscript{\rm 1\thanks{Corresponding author.}}, Chi Harold Liu\textsuperscript{\rm 1}, Xinjing Cheng\textsuperscript{\rm 2,\rm 3}
}
\affiliations{
    \textsuperscript{\rm 1}School of Computer Science and Technology, Beijing Institute of Technology, Beijing, China \\
    \textsuperscript{\rm 2}School of Software, BNRist, Tsinghua University, Beijing, China \\ 
    \textsuperscript{\rm 3}Inceptio Technology, Shanghai, China \\

    \{mingjiali, binhuixie, shuangli, chiliu\}@bit.edu.cn, cnorbot@gmail.com
}

\begin{document}

\maketitle

\begin{abstract}
    Generalizing models trained on normal visual conditions to target domains under adverse conditions is demanding in the practical systems. One prevalent solution is to bridge the domain gap between clear- and adverse-condition images to make satisfactory prediction on the target. However, previous methods often reckon on additional reference images of the same scenes taken from normal conditions, which are quite tough to collect in reality. Furthermore, most of them mainly focus on individual adverse condition such as nighttime or foggy, weakening the model versatility when encountering other adverse weathers. To overcome the above limitations, we propose a novel framework, Visibility Boosting and Logit-Constraint learning (\texttt{VBLC}), tailored for superior normal-to-adverse adaptation. \texttt{VBLC} explores the potential of getting rid of reference images and resolving the mixture of adverse conditions simultaneously. In detail, we first propose the \boostModuleName~to dynamically improve target images via certain priors in the image level. Then, we figure out the overconfident drawback in the conventional cross-entropy loss for self-training method and devise the \lossModuleName, which enforces a constraint on logit outputs during training to mitigate this pain point. To the best of our knowledge, this is a new perspective for tackling such a challenging task. Extensive experiments on two normal-to-adverse domain adaptation benchmarks, i.e., Cityscapes $\to$ ACDC and Cityscapes $\to$ FoggyCityscapes + RainCityscapes, verify the effectiveness of \methodblank, where it establishes the new state of the art. Code is available at \url{https://github.com/BIT-DA/VBLC}.
\end{abstract}

\section{Introduction}

\begin{figure}[t]
    \centering
    \includegraphics[width=0.45\textwidth]{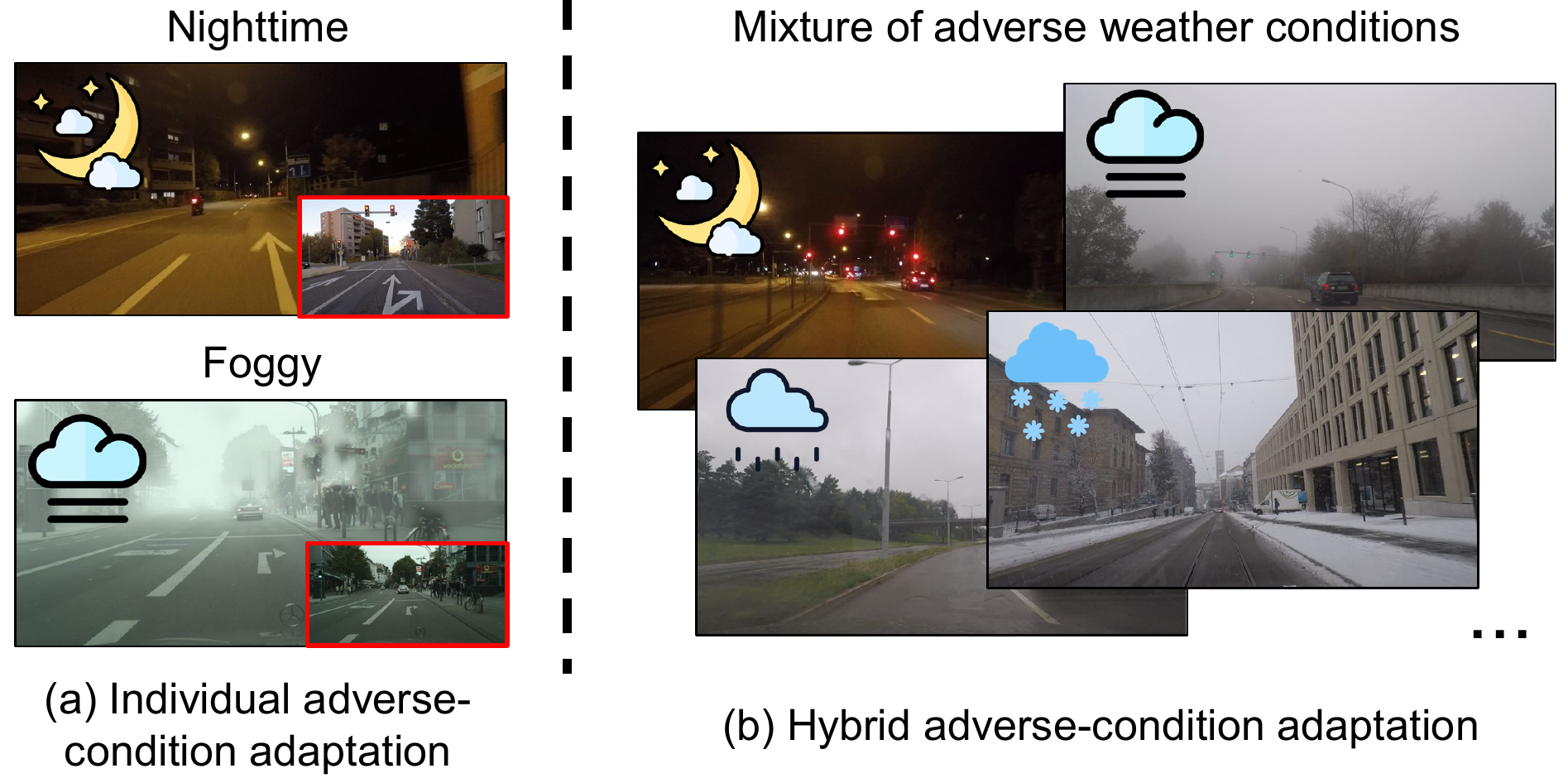}
    \caption{{\bf Problem comparison.} (a) Individual adverse-condition adaptation: reference images (shown in the red box at bottom right corner) depicting a similar scene are leveraged as an intermediate domain to assist in a specific adverse condition, e.g., nighttime or foggy. (b) Hybrid adverse-condition adaptation: the mixture of images from multiple adverse conditions are used without any reference.}
    \label{fig:motivation}
\end{figure}

The past few years have witnessed predominance of deep learning based methods in fundamental vision tasks, where scene understanding under extreme vision conditions has been attracting substantial research interest~\cite{ma2022both,sakaridis2020map}. 
In many outdoor applications, adverse weather conditions are frequently encountered, causing poor visibility and performance degradation.
For a safer, smoother operating environment, a desirable perception system should be trustworthy under a wide variety of scenarios~\cite{zhang2021autonomous,sakaridis2021acdc}. But, existing studies are mostly centered around datasets consisting of clean images, yet ignore the challenge of driving in varying adverse weather conditions, making them vulnerable in practice. Meanwhile, it is implausible to collect a dataset that fully reflects all situations and then separate data into discrete domains, since the visual appearance changes overtime and depends on the specific location, season and many other factors, all of which introduce a natural domain shift between any training (source) and test (target) distributions.

Accordingly, the emergence of more robust models for adverse conditions is vital to paving the way for their real-world utility. Unsupervised domain adaptation (UDA) is an alternative method~\cite{ganin2015dann_jmlr,tzeng2017adversarial,long2018conditional,liu2020ocda,ParkWSK20,GDCAN} to adapt models trained with well-labeled clear (source) images to adverse (target) images without access to target annotations. Until now, nighttime image segmentation and foggy scene understanding are two mainstream tasks. Given the difficulties of both specific problems, a great deal of works, such as~\cite{sakaridis2020map,DaiSHG20,wu2021one,ma2022both}, are carefully designed and highly customized, with the significant prior knowledge, e.g., additional clear-condition images. In Fig.~\ref{fig:motivation}(a), reference images (red boxes), depicting the similar scenes in correspondence with adverse images, are meant to boost segmentation performance. Unfortunately, it is no picnic to gather exactly paired images in the rapidly changing driving scenes. On the other hand, such a clear and specific distinction among adverse conditions is hard to define, e.g., test images are continually varying, which could be collected in composite conditions.

Driven by the above analysis, we advocate a new approach without the need of extra clear images for reference, dubbed as Visibility Boosting and Logit-Constraint learning (\texttt{VBLC}). It's worth noting that, this setting is generally regarded as under-constrained, making it quite difficult and rarely researched into. Take it a step further, we concentrate on a much more practical scenario where input images feature a hybrid of adverse conditions, i.e., low-light and flare characteristics of nighttime, veiling effects formed by heavy rain, dense foggy, snow, and so on (see Fig.~\ref{fig:motivation}(b)). 

To begin with, we introduce the \boostModuleName~in the input space to close the gap between normal- and adverse-condition images without the reliant on normal-adverse image pairs for reference. The absence of such weak supervision urges us to make the most of the priors as a replacement. We provide a saturation-based prior to adaptively heighten the visibility of incoming images. On top of that, boosted images are incorporated during training to bridge the immense gap brought about by adverse conditions. 

Second, for the self-training schemes prevailing in UDA, we observe the insufficient exploitation of predictions on unlabeled target samples for fear of overconfidence~\cite{wei2022mitigating}. To resolve this, we further come up with the \lossModuleName~to relieve the stringent demand on the quality of pseudo labels. Through gradient analysis, the constraint on the logit outputs during training can slow down the trend towards overconfidence and capitalize on predictions.  

Eventually, we show that \method establishes state-of-the-art performance on two challenging benchmarks. Compared to the current SOTA method, \method improves the relative performance by 8.9\% and 4.9\% (mIoU) on Cityscapes $\to$ ACDC and Cityscapes $\to$ FoggyCityscapes + RainCityscapes, respectively. We summarize contributions below:
\begin{itemize}
    \item We tackle a more realistic and challenging task of domain adaptive semantic segmentation under adverse conditions without the aid of extra image counterparts to form a clear-adverse pair.
    \item We desire to fill the blank through making adjustments at both ends of the network. The \boostModuleName~is proposed to narrow the visibility gap in the input space, while the \lossModuleName~is included in the output space to handle the overconfidence issue.
    \item We justify the effectiveness of our \method and explore the mechanism behind its success via extensive experiments.
\end{itemize}

\section{Related Work}\label{sec:related}

\paragraph{Normal-to-Adverse Domain Adaptation.} Domain adaptation has been well investigated in both theory~\cite{ben2010theory} and practice~\cite{WangD18}. Here, we are particularly interested in semantic segmentation task. \textit{Adversarial training} is the most examined method that narrows the domain gap via style transfer~\cite{hoffman2018cycada} or learning indistinguishable representations~\cite{tsai2018learning,vu2019advent,kim2020learning}. Recently, \textit{self-training} methods turn to pseudo labels to acquire extra supervision, reaching better performance. Advanced practices are to improve the quality of pseudo labels~\cite{zou2018unsupervised,zou2019confidence}, stabilize the training process~\cite{tranheden2021dacs,lukas2021daformer}, or utilize auxiliary task~\cite{wang2021domain,xie2022sepico}.

Despite the rising interest in developing domain adaptation models, existing works mostly concentrate on handling domain shifts introduced by the limitations of scene synthesis or by visual differences due to the variation in shooting locations. Considerably fewer attempts have been made to mitigate the shifts posed by adverse conditions, which is especially critical in reality~\cite{sakaridis2021acdc,liu2022image}.
Equipped with abundant data from various domains, several methods resort to curriculum-based schemes to realize a progressive adaptation towards a distant target domain~\cite{wulfmeier2018incremental,sakaridis2018model}. The dilemma of this scheme is that manually assigned intermediate domains may be suboptimal or arduous to design. Another promising direction is to make the best of the corresponding image pairs in dataset.
\citet{ma2022both} decouple style factor, fog factor and dual factor to cumulatively adapt these three factors. Alternatively, pixel-level warping is employed to benefit the prediction of static classes~\cite{wu2021one}, enable multi-view prediction fusion~\cite{sakaridis2020map}, or guide the subsequent label correction~\cite{bruggemann2022refign}. 

In general, the above methods require corresponding clear images, while the setting excluding image correspondences has rarely been explored. In this work, we are capable of addressing arbitrary adverse conditions without leveraging such weak supervision for adaptation.

\paragraph{Multi-Target Domain Adaptation/Generalization.} The goal is to extend domain adaptation to multiple target domains, which is relevant to our work.
Works in this field usually employ domain transfer~\cite{lee2022adas}, knowledge distillation~\cite{isobe2021multi}, curriculum learning~\cite{liu2020ocda} or meta-learning~\cite{GongCPLCLDG21} to bootstrap generalization across domains. To name a few, \citet{ParkWSK20} decompose a hard problem into multiple easy single-target adaptation problems. 
\citet{lee2022adas} perform style transfer in the input space and utilize a direct adaptation strategy towards multiple domains.  
Our approach differs from these methods in the way we treat the target samples, where images under widely varied adverse scenarios are viewed as a mixture of multiple domains, and domain labels in the test set are unavailable.

\begin{figure*}[t]
    \centering
    \includegraphics[width=0.96\textwidth]{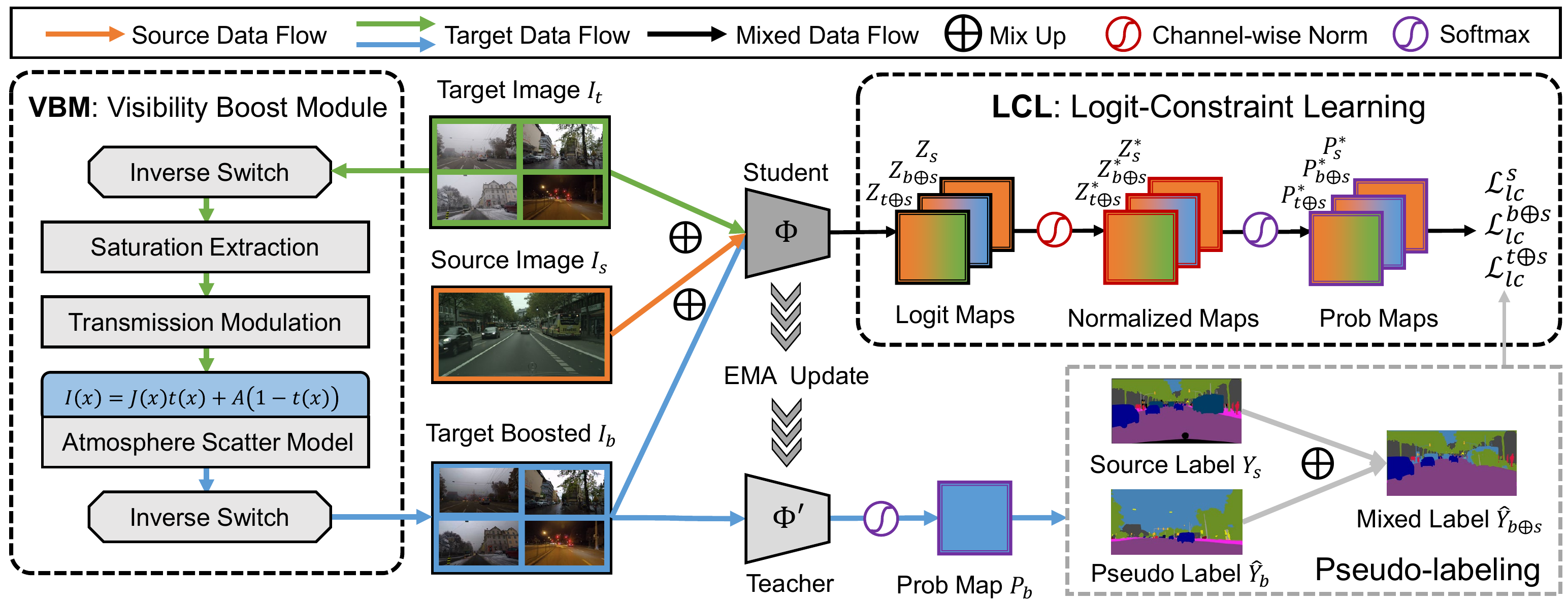}
    \caption{{\bf Overview of \methodblank}. Our framework enhances the capability of self-training schemes at both ends of the pipeline. In the input space, the \boostModuleName~is incorporated to ameliorate target images and generate more reliable pseudo labels. In the output space, the specialized \lossModuleName~is devised to conquer the erroneous prediction brought about by tremendous domain gap. Together with slight modifications to the training scheme, a simple, competitive approach is proposed.}
    \label{fig:framework}
\end{figure*}

\paragraph{Poor Visibility Image Enhancement.} There is a significant body of works whose goal is increasing image visibility.
Research interest in the mechanism of haze formation~\cite{mccartney1976optics} has emerged long before the neural nets prevail. Since then, numerous conditions featuring poor visibility are described including low-light~\cite{land1977retinex}, fog~\cite{narasimhan2003contrast}, rain~\cite{li2019heavy}, and snow~\cite{chen2020jstasr}. 
Subsequent deep learning-based works either alleviate the burden of hyperparameter tuning via a neural net~\cite{liu2022image}, or leave priori factors completely behind and rely on neural nets to automatically adapt to different physical models~\cite{valanarasu2022transweather}. Despite these efforts, they always need paired images that are hard or even impossible to acquire due to the dynamic scenes in reality, to cater to network optimization. By contrast, we only utilize the single image with poor visibility and manage to design a dynamic module as a substitute for the manual tuning process.

\section{Method}\label{sec:method}

Given source images with pixel-level annotations from a normal-condition source domain $\mathcal{S}$ (e.g., good weather, favorable illumination), and unlabeled target images from an adverse-condition target domain $\mathcal{T}$ (e.g., nighttime, fog, rain, snow, etc.), the goal is to predict high-quality segmentation maps for the target domain. Note that the target domain described above could turn out to be a combination of several domains, but we blur their boundaries and regard them as featuring diversity in a single target domain.

The full pipeline of our method is illustrated in Fig.~\ref{fig:framework}. Overall, it contains two major parts: (i) a basic framework that is composed of a teacher model for pseudo-labeling and a student model for online learning; (ii) two dedicated modules which encourage reducing domain differences in imaging conditions and output configurations. In the following, we provide a detailed description of \boostModuleName~as well as \lossModuleName. After that, the overall optimization and algorithm are introduced.

\subsection{\boostModuleNameTitle}
Previous methods generally utilize the normal-adverse image correspondences, which may put stress on data collection and annotation. Here, we merely exploit the images under adverse conditions in the target domain.
At first, an appropriate prior should come in place to break out of the dilemma, and the atmosphere scatter model (ASM)~\cite{nayar1999vision,narasimhan2003contrast} is a powerful candidate to describe haze formation:
\begin{small}
    \begin{equation}
        \begin{aligned}
            I(x) = J(x)t(x) + A(1-t(x))\,,
            \label{eq:asm}
        \end{aligned}
    \end{equation}
\end{small}%
where $I(x)$ is the observed hazy image, $J(x)$ is the scene radiance, namely restored image, and $A$ is the atmospheric light estimated globally. $t(x)$ represents the transmission map describing the portion of light that survives scattering and reaches the camera. It is represented as: $t(x) = e^{-\beta d(x)}$, where $\beta$ is the scattering coefficient and $d(x)$ denotes pixel-wise scene depth. As depicted in \cite{li2019heavy}, the vanilla ASM can already model the veiling effect usually observed in fog, heavy rain, or even snowy scenes. 

We consider such effect as the major obstacle lying in the way to a clear vision, and intend to alleviate the problem in an adaptive way. 
Motivated by~\cite{he2009single,liu2022image}, we propose the \boostModuleName~(VBM) to ameliorate images in various adverse conditions. We estimate the atmospheric light $A$ from the 1,000 brightest pixels in the image, and the transmission map can be given as:
\begin{small}
    \begin{equation}
        \begin{aligned}
            t(x) = 1 - \min\limits_{c}\left(\min\limits_{y\in\Omega(x)} \frac{I^c(y)}{A^c}\right)\,,
        \end{aligned}
    \end{equation}
\end{small}%
where $c\in\{r,g,b\}$ is the color channel and $\Omega(x)$ is the local patch surrounding position $x$.
To make the restored image more natural in appearance, we further devise a non-parametric coefficient $\omega_s$ to control the dehaze extent. Hereafter, the $t(x)$ is reformulated as
\begin{small}
    \begin{equation}
        \begin{aligned}
            t(x) = 1 - \omega_s\min\limits_{c}\left(\min\limits_{y\in\Omega(x)} \frac{I^c(y)}{A^c}\right)\,,
            \label{eq:tx}
        \end{aligned}
    \end{equation}
\end{small}%
where $\omega_s$ is an adaptive coefficient for transmission modulation. Through observation, it can be summarized that the images with veiling effect tend to be gray and dull, which can be approximately described as low saturation. On the contrary, the images with clear vision can be more vivid and colorful, resulting in greater saturation. By introducing this coefficient, the restored appearance can be constrained to some extent. To be precise, $\omega_s$ is dynamically calculated from the mean value of saturation ${mean}_s$ within an image:
\begin{equation}
    \begin{aligned}
        \omega_s = e^{- {mean}_{s}\times \gamma}\,,
        \label{eq:omega}
    \end{aligned}
\end{equation}%
where $\gamma$ is a scaling factor that is experimentally fixed to $4.0$. And ${mean}_s$ is calculated by:
\begin{small}
    \begin{equation}
        \begin{aligned}
            {mean}_s=\frac{1}{HW}\sum_{h, w}\frac{\max\limits_{c}I_{h, w}^c-\min\limits_{c}I_{h, w}^c}{\max\limits_{c}I_{h, w}^c}\,,
        \end{aligned}
    \end{equation}
\end{small}%
where $H, W$ are height and width of the image $I$, and $I_{h, w}^c$ is the $c^{th}$ color channel of the pixel indexed $(h, w)$ in $I$. 

However, one exception is the nighttime condition. Luckily, as claimed by~\citet{zhang2012enhancement}, the reverted low-light images can be viewed as hazy images. Thus, we additionally add a pair of \textit{Inverse Switch} to extend VBM to low-light condition, by which only night images are inverted and other types of images remain unchanged. We formulate this as:
\begin{small}
    \begin{equation}
        \begin{aligned}
            1-I(x) = (1-J(x))t(x) + A(1-t(x))\,.
        \end{aligned}
    \end{equation}
\end{small}%
With conditional inversion, one can handle all cases without modifying the core procedure of visibility enhancement. Note that Eq.~\eqref{eq:tx} may seem close to the ones proposed in~\cite{he2009single} and~\cite{liu2022image}, but is different as it is neither manually tuned for each image nor reliant on paired image to learn a parameter. Actually, the scale factor $\gamma$ is globally assigned, and then the coefficient $\omega_s$ can dynamically adapt to different images.

The complete pipeline of VBM is illustrated in the left of Fig.~\ref{fig:framework}. For an arbitrary adverse-condition image, we first decide the application of \textit{Inverse Switch} according to its lighting condition, and the mean saturation value ${mean}_s$ is extracted form the original/inverted image. After that, the coefficient $\omega_s$ is yielded by Eq.~\eqref{eq:omega}, which is then used to perform transmission modulation in Eq.~\eqref{eq:tx}. The image is then enhanced by ASM. Another \textit{Inverse Switch} is applied in parallel with the aforementioned one. Through the above process, visibility enhanced target images can be obtained. 

\subsection{\lossModuleNameTitle}
\label{sec:loss}

In the literature, it is almost common practice to combine the self-training strategy with cross-entropy (CE) loss for both source and target samples~\cite{zou2019confidence,zou2018unsupervised,tranheden2021dacs,xie2022sepico}. For simplicity, we take a pixel as an example, whose CE loss is formulated as follows:
\begin{small}
    \begin{equation}
        \begin{aligned}
            \lossce = -\sum\limits_{k=1}^K y_k\log(p_k),~\text{where}~~p_i = \frac{e^{z_i}}{\sum_{k=1}^K e^{z_k}}\,,
            \label{eq:loss_ce}
        \end{aligned}
    \end{equation}
\end{small}%
where $K$ represents the number of classes, $y$ is a one-hot ground-truth (pseudo) label, $p_i$ is the probability for the $i^{th}$ class, and $z_i$ is the $i^{th}$ element of the logit output.

As we all known, the CE loss forces predictions to resemble the corresponding one-hot labels, which makes it effective for supervised training paradigms. However, when it comes to unlabeled target samples, this characteristic can be a mixed blessing: reliable pseudo labels can compensate the deficiency of ground-truth labels, but erroneous pseudo labels can be catastrophic as a strong supervision. Existing self-training methods are well aware of such risk, and have attempted to address the issue through loss reweighting~\cite{olsson2021classmix,tranheden2021dacs}, confidence thresholding~\cite{he2021re}, or pseudo label refinement~\cite{zhang2021prototypical}. Nevertheless, these methods either generate pseudo labels regardless of their confidence, or just ignore the pixels under the threshold of confidence policy, failing to make the most of precious predictions. Furthermore, in the task of normal-to-adverse adaptation, trustworthy predictions can be rather scarce, thus blindly ignoring the unconfident pixels will result in low data efficiency.

To address this issue, we seek to push the utilization of model prediction to a new height through the enhancement of loss term, for its close relation with predictions. As the unconfident samples are promising providers of extra information, we contend that inter-class relationship within the prediction of a pixel should be emphasized. When taking the derivability into consideration, $\ell_2$-norm is an ideal candidate as it does link all elements in an equal manner. Therefore, we integrate the $\ell_2$-norm into the original CE loss as an expansion, forming a new \lossModuleName~loss:
\begin{small}
    \begin{equation}
        \begin{aligned}
            \loss = -\sum\limits_{k=1}^K y_k\log(p_k^*),~\text{where}~~p_i^* = \frac{e^{z_i / {\Vert z \Vert}}}{\sum_{k=1}^K e^{z_k / {\Vert z \Vert}}} \,,
            \label{eq:loss_md}
        \end{aligned}
    \end{equation}
\end{small}%
where $\Vert\cdot\Vert$ means $\ell_2$-norm. The name \lossModuleName~comes from the fact that every logit element is rescaled by dividing the norm term, whose optimization is thus constrained by the other elements constituting the logit. We will theoretically reveal the inter-class constraint through gradient analysis in the following part.

\paragraph{Gradient Analysis.} It's worth mentioning that this new loss function is not confined to the confident portion of predictions, but can be applied to unconfident predictions with a unified form. To justify our claim, let's take a closer look at the gradient during back-propagation. For the vanilla CE loss, the gradient of loss to the $j^{th}$ logit element is:
\begin{small}
    \begin{equation}
        \begin{aligned}
        \frac{\partial \mathcal{L}_{ce}}{\partial z_j} = p_j - y_j\,.
        \label{eq:ce_grad}
        \end{aligned}
    \end{equation}
\end{small}%
Just as discussed above, this gradient merely narrows the gap between the prediction and the corresponding label, making it inevitable to ruin the prediction if a wrong label is given. On the contrary, the gradient of our proposed \lossModuleName~loss to the $j^{th}$ logit element is:
\begin{small}
    \begin{equation}
        \begin{aligned}
        \frac{\partial \loss}{\partial z_j} = \frac{1}{\Vert z \Vert}\left((p_j^* - y_j) - \sum\limits_{k=1}^K \frac{z_j z_k}{{\Vert z \Vert}^2}(p_k^* - y_k)\right)\,.
        \label{eq:norm_grad}
        \end{aligned}
    \end{equation}
\end{small}%
In this formula, the gradient is made up of two parts. The former part is essentially identical to the gradient of vanilla CE loss, while the latter part undoubtedly reflect the connection built across different classes. More derivations can be found in the Appendix~\ref{sec:proof}.

Let us analysis the gradient from both confident and unconfident conditions. Assuming the prediction is confident, then the coefficient of the second term, namely $\frac{z_j}{\Vert z\Vert}*\frac{z_k}{\Vert z\Vert}$, should be relative small except for $k=j$, and the gradient can be approximate to
\begin{small}
    \begin{equation}
        \begin{aligned}
            \frac{1}{\Vert z\Vert}\left((p_j^* - y_j) - (\frac{z_j}{\Vert z\Vert})^2(p_k^* - y_k)\right) \,.
        \end{aligned}
    \end{equation}
\end{small}%
If there is space left for optimization, i.e., $z_j < \Vert z\Vert$, this term is still capable of providing gradient; otherwise, the gradient degrades to zero and the optimization stops. When confronted with unconfident predictions, the coefficient of the second term would be relatively larger for all classes whose prediction $p_k$ is close to that of $p_j$, given that $z$ is directly related to $p$, thus reducing the gradient and slowing down the optimization towards the assigned pseudo label. 

In a nutshell, the above gradient analysis not only reflects the ability of \lossModuleName~to follow the guidance of confident pseudo label, but highlights its potential to explore the knowledge hidden in unconfident predictions without fear of overconfidence. More experimental analysis can be found in Section~\ref{sec:ablation}.

\subsection{Overall Optimization}
During the training stage, images from both source and target domains are first randomly sampled, i.e., $\sourceimage, \targetimage\in R^{H\times W \times 3}$, respectively. The target image is then passed into \boostModuleName~to obtain boosted target image $\boostimage$ with better visibility. Subsequently, the pseudo label $\targetpseudo$ of $\boostimage$ is predicted from the teacher model $\Phi'$. Next,  both original target image and boosted one are separately mixed up with the source image using Classmix~\cite{olsson2021classmix} for online learning. The blended images are noted as $\mixtargetimage$ and $\mixboostimage$, which are then passed through the student model $\Phi$ to get logit outputs $\mixtargetlogit$ and $\mixboostlogit$, respectively. And they share the same mixed label $\mixlabel$ that is mixed up between source ground-truth label $\sourcelabel$ and target pseudo label $\targetpseudo$. Before participating in the final loss calculation, logits are processed through channel-wise norm and softmax function to get the final prediction maps $\sourceprob,\mixtargetprob, \mixboostprob$. Eventually, for any image, the \lossModuleName~loss is given by:
\begin{small}
    \begin{equation}
            \loss(Y, P^*) = - \frac{1}{HW}\sum\limits_{h,w}\sum\limits_{k=1}^{K} {Y_{h,w,k} \log{P_{h,w,k}^*}},
        \label{eq:loss_md_all_pixel}
    \end{equation}
\end{small}%
where $H, W$ are height and width of an input image, $Y_{h,w,k}$ is the $k^{th}$ element in a one-hot label of pixel indexed $(h,w)$, and $P_{h,w,k}^*$ is the corresponding pixel-level prediction.

For the source data, we have $\loss^s = \loss(Y_s, \sourceprob)$. For the target data, a quality estimation is produced for the pseudo labels following \citet{tranheden2021dacs}. Here an adaptive weight $\lambda_\delta$ is calculated from the proportion of confident  pixel-level predictions ($\max_k P_{h,w,k}^*>\delta$) and weighted on losses involving target images. Subsequently, the losses for blended images $\mixtargetimage, \mixboostimage$ can be respectively computed as $\loss^{t\oplus s} = \lambda_\delta \loss(\mixlabel, \mixtargetprob)$ and $\loss^{b\oplus s} = \lambda_\delta \loss(\mixlabel, \mixboostprob)$.
Overall, the training objective can be formulated as:
\begin{small}
    \begin{equation}
        \begin{aligned}
            \min_\Phi~\loss^s + \loss^{t\oplus s} + \loss^{b\oplus s}.
            \label{eq:overall_loss}
        \end{aligned}
    \end{equation}
\end{small}%
By default, loss weighting coefficients are set to 1.0. For a detailed schedule, please refer to Alg.~\ref{alg:algorithm} in the Appendix~\ref{sec:alg}.

\begin{table*}
  \centering
  \resizebox{0.99\textwidth}{!}{
    \def\arraystretch{1.1}
    \begin{tabular}{ l | c c c c c c c c c c c c c c c c c c c | c }
        \Xhline{1.2pt}
        Method & \rotatebox{0}{road} & \rotatebox{0}{side.} & \rotatebox{0}{buil.} & \rotatebox{0}{wall} & \rotatebox{0}{fence} & \rotatebox{0}{pole} & \rotatebox{0}{light} & \rotatebox{0}{sign} & \rotatebox{0}{veg.} & \rotatebox{0}{terr.} & \rotatebox{0}{sky} & \rotatebox{0}{pers.} & \rotatebox{0}{rider} & \rotatebox{0}{car}& \rotatebox{0}{truck} & \rotatebox{0}{bus} & \rotatebox{0}{train} & \rotatebox{0}{mbike} & \rotatebox{0}{bike} & mIoU \\
        \hline
        \hline
        DeepLab-v2 & 71.9 & 26.2 & 51.1 & 18.8 & 22.5 & 19.7 & 33.0 & 27.7 & 67.9 & 28.6 & 44.2 & 43.1 & 22.1 & 71.2 & 29.8 & 33.3 & 48.4 & 26.2 & 35.8 & 38.0 \\
        AdaptSegNet & 69.4 & 34.0 & 52.8 & 13.5 & 18.0 & 4.3 & 14.9 & 9.7 & 64.0 & 23.1 & 38.2 & 38.6 & 20.1 & 59.3 & 35.6 & 30.6 & 53.9 & 19.8 & 33.9 & 33.4 \\
        ADVENT & 72.9 & 14.3 & 40.5 & 16.6 & 21.2 & 9.3 & 17.4 & 21.2 & 63.8 & 23.8 & 18.3 & 32.6 & 19.5 & 69.5 & 36.2 & 34.5 & 46.2 & 26.9 & 36.1 & 32.7 \\
        BDL & 56.0 & 32.5 & 68.1 & 20.1 & 17.4 & 15.8 & 30.2 & 28.7 & 59.9 & 25.3 & 37.7 & 28.7 & 25.5 & 70.2 & 39.6 & 40.5 & 52.7 & 29.2 & 38.4 & 37.7 \\
        CLAN & \textbf{79.1} & 29.5 & 45.9 & 18.1 & 21.3 & 22.1 & 35.3 & 40.7 & 67.4 & 29.4 & 32.8 & 42.7 & 18.5 & 73.6 & 42.0 & 31.6 & 55.7 & 25.4 & 30.7 & 39.0 \\
        CRST & 51.7 & 24.4 & 67.8 & 13.3 & 9.7 & 30.2 & 38.2 & 34.1 & 58.0 & 25.2 & \textbf{76.8} & 39.9 & 17.1 & 65.4 & 3.7 & 6.6 & 39.6 & 11.8 & 8.6 & 32.8 \\
        FDA & 73.2 & 34.7 & 59.0 & 24.8 & \textbf{29.5} & 28.6 & 43.3 & 44.9 & \textbf{70.1} & 28.2 & 54.7 & 47.0 & 28.5 & 74.6 & \textbf{44.8} & \textbf{52.3} & \textbf{63.3} & 28.3 & 39.5 & 45.7 \\
        DACS & 58.5 & 34.7 & 76.4 & 20.9 & 22.6 & 31.7 & 32.7 & 46.8 & 58.7 & 39.0 & 36.3 & 43.7 & 20.5 & 72.3 & 39.6 & 34.8 & 51.1 & 24.6 & 38.2 & 41.2 \\
        \gr \bf \method & 49.6 & \textbf{39.3} & \textbf{79.4} & \textbf{35.8} & \textbf{29.5} & \textbf{42.6} & \textbf{57.2} & \textbf{57.5} & 69.1 & \textbf{42.7} & 39.8 & \textbf{54.5} & \textbf{29.3} & \textbf{77.8} & 43.0 & 36.2 & 32.7 & \textbf{38.7} & \textbf{53.4} & \textbf{47.8} \\
        \hline
        SegFormer & 66.9 & 25.8 & 71.3 & 20.9 & 22.2 & 41.1 & 47.2 & 46.6 & \textbf{74.2} & 44.9 & 75.6 & 50.4 & 23.5 & 73.1 & 30.3 & 36.8 & 55.8 & 29.4 & 37.1 & 45.9 \\
        DAFormer &  56.9 & 45.4 & 84.7 & \textbf{44.7} & 35.1 & 48.6 & 44.8 & 57.4 & 69.5 & 52.9 & 45.8 & 57.1 & 28.2 & 82.8 & 57.2 & 63.9 & 84.0 & 40.2 & 50.5 & 55.3 \\
        \gr \bf\method & \textbf{89.2} & \textbf{59.8} & \textbf{85.9} & 44.0 & \textbf{37.2} & \textbf{53.5} & \textbf{64.5} & \textbf{63.2} & 72.4 & \textbf{56.3} & \textbf{84.1} & \textbf{65.5} & \textbf{37.7} & \textbf{85.1} & \textbf{60.1} & \textbf{71.8} & \textbf{85.2} & \textbf{47.7} & \textbf{56.3} & \textbf{64.2} \\
        \Xhline{1.2pt}
    \end{tabular}
  }

  \caption{Comparison with the state-of-the-arts on \textbf{Cityscapes $\rightarrow$ ACDC semantic segmentation task}. IoU score of each class and the mIoU score are reported on ACDC testing set. The bests results are highlighted in \textbf{bold}.
  } \label{table:acdc_class}
\end{table*}

\begin{table*}
  \centering
  \resizebox{0.99\textwidth}{!}{
    \def\arraystretch{1.1}
    \begin{tabular}{ l | c c c c c c c c c c c c c c c c c c c | c }
        \Xhline{1.2pt}
        Method & \rotatebox{0}{road} & \rotatebox{0}{side.} & \rotatebox{0}{buil.} & \rotatebox{0}{wall} & \rotatebox{0}{fence} & \rotatebox{0}{pole} & \rotatebox{0}{light} & \rotatebox{0}{sign} & \rotatebox{0}{veg.} & \rotatebox{0}{terr.} & \rotatebox{0}{sky} & \rotatebox{0}{pers.} & \rotatebox{0}{rider} & \rotatebox{0}{car}& \rotatebox{0}{truck} & \rotatebox{0}{bus} & \rotatebox{0}{train} & \rotatebox{0}{mbike} & \rotatebox{0}{bike} & mIoU \\
        \hline
        \hline
        DeepLab-v2 & 96.7 & 72.4 & 74.1 & 28.6 & 41.4 & 42.2 & 49.8 & 67.6 & 72.6 & 62.5 & 80.6 & 70.4 & 54.4 & 88.4 & 56.1 & 72.4 & 33.7 & 42.7 & 70.1 & 61.9 \\
        FDA & 87.0 & 56.9 & 82.1 & 4.3 & 11.6 & 36.3 & 41.8 & 60.4 & 80.6 & 51.6 & 70.6 & 66.7 & 50.3 & 86.0 & 46.4 & 63.7 & 26.2 & 41.4 & 66.3 & 54.2 \\
        DACS & 97.9 & 82.3 & \textbf{88.7} & 40.8 & 42.4 & 41.0 & 53.5 & 67.3 & \textbf{89.2} & 58.2 & \textbf{90.8} & 70.8 & 54.4 & 91.3 & 62.9 & 82.5 & \textbf{56.4} & 47.0 & 72.4 & 67.9 \\
        \gr \bf \method & \textbf{98.6} & \textbf{86.9} & 87.2 & \textbf{62.1} & \textbf{55.3} & \textbf{54.2} & \textbf{65.1} & \textbf{77.8} & 86.9 & \textbf{66.8} & 90.1 & \textbf{77.5} & \textbf{63.2} & \textbf{93.7} & \textbf{77.3} & \textbf{86.6} & 55.0 & \textbf{59.4} & \textbf{79.5} & \textbf{74.9} \\
        \hline
        SegFormer & 97.8 & 81.6 & 86.9 & 54.3 & 48.3 & 49.2 & 57.3 & 71.6 & 86.9 & 65.5 & 83.4 & 71.9 & 57.1 & 91.8 & 67.9 & 80.1 & 73.1 & 49.9 & 74.6 & 71.0 \\
        DAFormer & 98.5 & 87.0 & 90.8 & 55.1 & 53.7 & 56.3 & 62.8 & 73.6 & 91.5 & 70.7 & 90.0 & 75.6 & 56.8 & 92.7 & 65.9 & 88.3 & 79.9 & 56.9 & 77.6 & 74.9 \\
        \gr \bf \method & \textbf{98.7} & \textbf{88.4} & \textbf{91.9} & \textbf{66.3} & \textbf{65.2} & \textbf{62.7} & \textbf{69.1} & \textbf{79.6} & \textbf{92.2} & \textbf{72.4} & \textbf{92.3} & \textbf{80.0} & \textbf{66.0} & \textbf{94.6} & \textbf{79.9} & \textbf{90.9} & \textbf{81.8} & \textbf{64.0} & \textbf{80.6} & \textbf{79.8} \\
        \Xhline{1.2pt}
      \end{tabular}
  }
  \caption{Comparison with the state-of-the-arts on \textbf{Cityscapes $\rightarrow$ FoggyCityscapes + RainCityscapes semantic segmentation task}. IoU score of each class and the mIoU score are reported. The bests results are highlighted in {\bf bold}.}
  \label{table:rainfog}
\end{table*}

\section{Experiments}\label{sec:Experiments}

In this section, we assess the effectiveness of \method under several adverse conditions. For each task, we first give a brief introduction to the datasets and architectures involved. Following up are experimental results and insight analyses. Limited by space, more details of dataset, implementation, and additional results are left for the Appendix.

\subsection{Normal-to-Adverse Domain Adaptation}

\paragraph{Datasets and Architectures.} We first take out the experiments on two challenging semantic segmentation tasks, i.e., Cityscapes~\cite{cordts2016cityscapes} $\rightarrow$ ACDC~\cite{sakaridis2021acdc} and Cityscapes~\cite{cordts2016cityscapes} $\rightarrow$ FoggyCityscapes~\cite{sakaridis2018semantic} + RainCityscapes~\cite{hu2019depth}. We further testify the generality of \method by performing object detection on the latter. Among these tasks, Cityscapes (source domain) serves as a collection of clear images, while images from other datasets (target domain) all feature degraded visibility to some extent. For this part, we experiment on both CNN-based DeepLab-v2~\cite{chen2017deeplab} and Transformer-based SegFormer~\cite{xie2021segformer} to give a whole picture of the segmentation quality of our method. As to object detection task, following~\citet{Wang2021SFA}, Deformable DETR~\cite{DeformableDETR} is adopted as the basic architecture.

For semantic segmentation task, we utilize per class Intersection-over-Union (IoU)~\cite{everingham2015IoU} and mean IoU (mIoU) over all classes as an evaluation. For object detection task, we report the standard average precision (AP) result under different IoU thresholds and object scales.

\begin{figure}[t]
    \centering
    \includegraphics[width=0.47\textwidth]{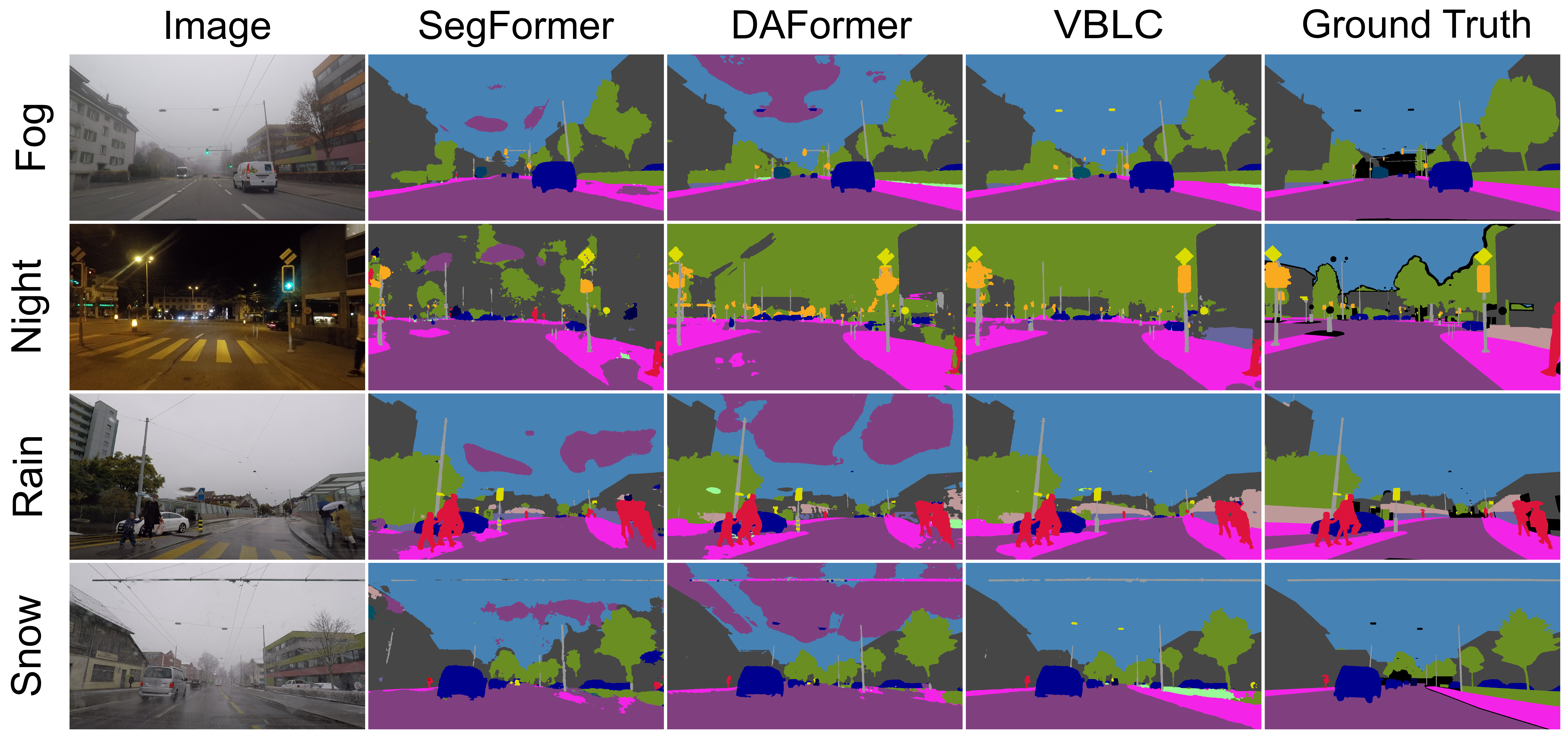}
    \caption{Visualization of segmentation results on ACDC validation set. From left to right: target images under distinct adverse conditions, results predicted by SegFormer, results predicted by DAFormer, and results predicted by our \method, ground-truth labels are shown one by one.}
    \label{fig:segmap}
\end{figure}

\paragraph{Experimental Results.} The comparison of our \method to relative methods on Cityscapes $\rightarrow$ ACDC segmentation task is listed in Table \ref{table:acdc_class}. Generally, the SegFormer-based methods substantially outperform the DeepLab-based ones. The previous state-of-the-art method built on DeepLab-v2 is FDA with a mIoU of 45.7\%, but our \method takes a step further and achieves 47.8\% mIoU, gaining a large boost of +2.1\% and could even rank among the SegFormer-based counterparts. When integrated with the stronger backbone, \method still yields a leading result of 64.2\% mIoU, outperforming DAFormer by a huge margin of +8.9\%. We also provide qualitative semantic segmentation results in Fig.~\ref{fig:segmap}. We can observe clear improvement against both Source-only and state-of-the-art adaptation (DAFormer) models, especially in the prediction of sky, light, and sign. 

Table \ref{table:rainfog} shows the segmentation results on Cityscapes $\rightarrow$ FoggyCityscapes + RainCityscapes. Due to slight domain shift, outcomes of the Source-only models are already high, however, our \method is still capable of providing consistent performance gain, surpassing DeepLab-v2 and SegFormer by +13.0\% mIoU and +8.8\% mIoU, respectively. This complementary experiment explores the robustness of our \method on tasks containing synthetic datasets, and proves the scalability of the proposed modules.

To showcase the flexibility of \methodblank, we further combine it with state-of-the-art UDA detection methods~\cite{Wang2021SFA} on Cityscapes $\to$ FoggyCityscapes + RainCityscapes object detection task. To be specific, images from the target domain is first boosted with the designed \boostModuleName, which performs coarse alignment between normal and adverse conditions. Then, the \lossModuleName~is integrated with class prediction.  And the results are reported in Table \ref{table:rainfog_det}. We can observe that \method boosts the performance of SFA by a substantial 1.3 AP, validating that our method can indeed generalize well under the variation of weather conditions, both for segmentation and detection.

\begin{table}[t]
  \centering
  \resizebox{0.47\textwidth}{!}{
    \def\arraystretch{1.1}
    \begin{tabular}{ l | c c cccc }
        \Xhline{1.2pt}
        Method & AP & AP$_{50}$ & AP$_{75}$ & AP$_{S}$ & AP$_{M}$ & AP$_{L}$ \\
        \hline
        \hline
        Deformable DETR & 13.4 & 22.7 & 13.4 & 3.4 & 17.0 & 26.8 \\
        SFA & 14.3 & 24.6 & 14.6 & 4.2 & 17.4 & 28.2 \\
        \gr SFA + \bf\method  & \bf 15.6 & \bf 26.0 & \bf 16.4 & \bf 5.2 & \bf 17.8 & \bf 30.9 \\
        \Xhline{1.2pt}
    \end{tabular}
  }
  \caption{Comparison with state-of-the-arts on \textbf{Cityscapes $\rightarrow$ FoggyCityscapes + RainCityscapes object detection task} with Deformable DETR~\cite{DeformableDETR}.}
  \label{table:rainfog_det}
\end{table}

\subsection{Multi-Target Domain Adaptation}
\begin{table}[t]
  \centering
  \resizebox{0.47\textwidth}{!}{
    \def\arraystretch{1.1}
    \begin{tabular}{ c | l | c c | c }
        \Xhline{1.2pt}
        \# Class & Method & IDD (mIoU) & Mapillary (mIoU) & Average mIoU \\
        \hline
        \hline
        \multirow{3}{*}{7}
        & MTKT & 68.3 & 69.3 & 68.8 \\
        & ADAS & 70.4 & \textbf{75.1} & 72.7 \\
        & \gc \bf\method & \gc \textbf{73.9} & \gc 71.7 & \gc \textbf{72.8} \\
        \hline
        \multirow{3}{*}{19}
        & CCL & \textbf{53.6} & 51.4 & 52.5 \\
        & ADAS & 48.3 & 53.6 & 50.5 \\
        & \gc \bf\method & \gc 52.9 & \gc \textbf{57.8} & \gc \textbf{55.3} \\
        \Xhline{1.2pt}
    \end{tabular}
  }
  \caption{Comparison with state-of-the-arts on \textbf{Cityscapes $\rightarrow$ IDD + Mapillary semantic segmentation task}. mIoU score of each domain and their average are reported.}
  \label{table:iddmap}
\end{table}

\paragraph{Datasets and Architectures.} Now we turn to multi-target domain adaptation (MTDA), adapting from Cityscapes to IDD~\cite{varma2019idd} and Mapillary~\cite{neuhold2017mapillary}. All datasets are captured in reality without specific inclusion of images under adverse conditions. We investigate this as a special case and compare our \method with other well-established MTDA methods. All methods mentioned in this part are built on DeepLab-v2 for a fair comparison. 

\paragraph{Experimental Results.} In accordance with previous attempts, we report the segmentation results on Cityscapes $\rightarrow$ IDD + Mapillary with both 19 classes and 7 super classes settings  in Table~\ref{table:iddmap}. On either of both, \method takes the lead regarding the average mIoU over two target domains, attaining 55.3\%/72.8\% mIoU for 19/7 classes, respectively. 
Note that \method is neither intended to deal with MTDA directly nor to enforce the dispersion of multiple target domains explicitly, it is still comparable to existing specially designed counterparts, which undoubtedly reflects its superiority.

\subsection{Ablation Studies}
\label{sec:ablation}

\paragraph{The Effect of Each Component on Cityscapes $\to$ ACDC.} We report the detailed improvements of each component in Table~\ref{table:ablation}. The first line presents the Source-only (SegFormer) model trained only on Cityscapes, which serves as the baseline with 45.9\% mIoU. When combined with conventional self-training on source-target blended image ($\losscetarget$), the performance is significantly boosted with a gain of +6.7\% mIoU, validating the huge potential of self-training scheme. 

Next, the \boostModuleName~(VBM) is integrated to ease the generation of pseudo labels. To be more precise, we directly utilize the prediction of boosted target image from teacher model as a guidance to both original and boosted target images. A moderate improve of performance can be witnessed in this process, and we attribute this phenomenon to the fact that prediction of boosted target images are appropriately constrained (VBM + $\lossceboost$). 

The penultimate line highlights the power of our \lossModuleName, which further brings a substantial increase of +6.2\% mIoU, leading to a competitive performance of 63.2\% mIoU ($\losstarget$ + $\lossboost$). Finally, we additionally apply the \lossModuleName~to the source domain, and obtain a bonus of +1.0\% mIoU ($\losssource$), yielding the ultimate score of 64.2\% mIoU. 
In summary, we can learn that \method mainly enhances the performance from two critical aspects, i.e., visibility boost and logit-constraint learning.

\begin{table}[t]
  \centering
  \resizebox{0.47\textwidth}{!}{
      \begin{tabular}{ c | c | c c c |c c c | c}
          \toprule[1.2pt]
          Method & VBM  & $\losscesource$ & $\losscetarget$ & $\lossceboost$ & $\losssource$ & $\losstarget$ & $\lossboost$ & mIoU \\
          \midrule
          Source-only & & \checkmark & & & & & & 45.9 \\
          \midrule
          \multirow{4}{*}{Ours} & & \checkmark & \checkmark & & & & & 52.6 (6.7$\uparrow$) \\
          & \checkmark & \checkmark & \checkmark & \checkmark & & & & 57.0 (4.4$\uparrow$) \\
          & \checkmark & \checkmark & & & & \checkmark & \checkmark & 63.2 (6.2$\uparrow$) \\
          & \checkmark & &  &  & \checkmark &  \checkmark & \checkmark & \bf{64.2} (1.0$\uparrow$) \\
          \bottomrule[1.2pt]
      \end{tabular}
  }
  \caption{Ablation study on \textbf{Cityscapes $\rightarrow$ ACDC semantic segmentation task}.}
  \label{table:ablation}
\end{table}

\begin{figure}[t]
    \centering
    \includegraphics[width=0.47\textwidth]{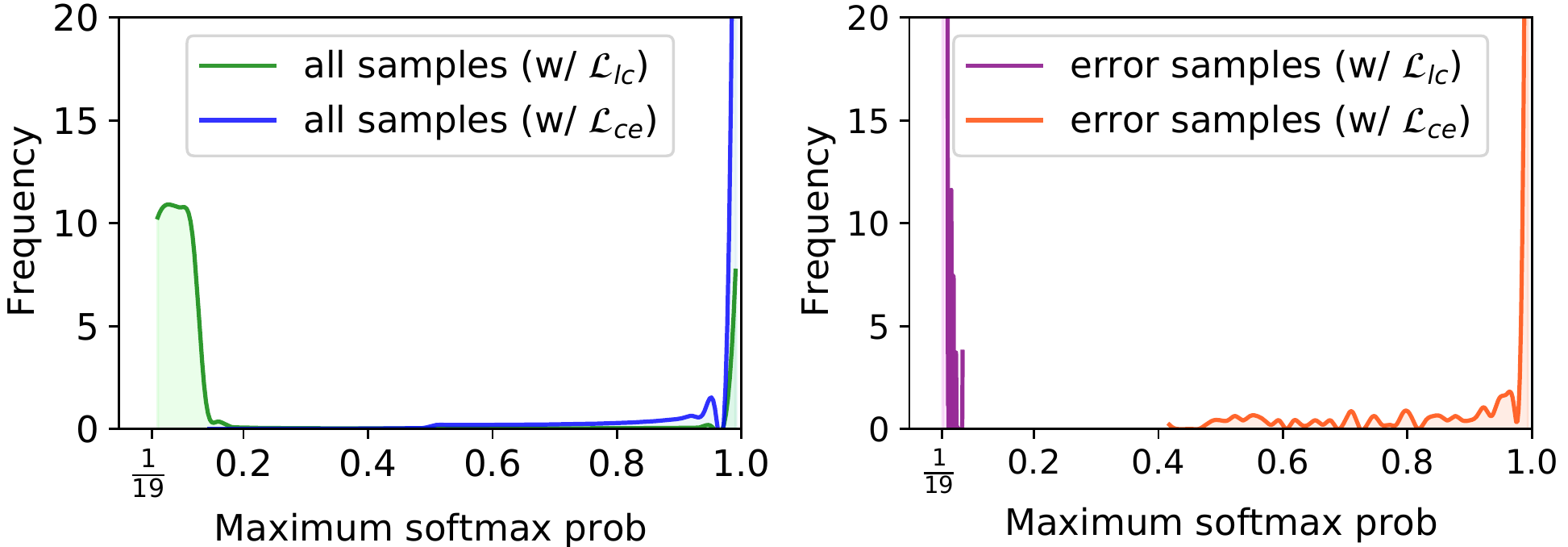}
    \caption{Confidence distribution over all (left chart) or erroneous (right chart) predictions on ACDC validation set.}
    \label{fig:conf}
\end{figure}

\paragraph{Analysis on $\loss$.} As mentioned in Section~\ref{sec:loss}, our \lossModuleName~is capable of constraining the optimization process to address the problem of overconfidence. Fig.~\ref{fig:conf} visualizes confidence distributions from models trained on vanilla CE loss ($\lossce$) and \lossModuleName~loss ($\loss$) separately on ACDC validation set. We opt to adopt max softmax as the confidence, without test-time logit constraint for a fair comparison. The left chart shows confidence distribution on the whole validation set. We can observe that, the model trained with $\lossce$ tend to be confident with a majority of predictions, while the one trained with $\loss$ remains skeptical to a handful of predictions. Indeed, the $\loss$ allows the coexistence of especially confident predictions and rather unconfident ones, which is in line with our analysis. The right chart, on the other hand, illustrates the confidence distribution on the erroneous predictions only. This chart can further reflect whether the predictions are overconfident. It is obvious that predictions from the model trained with $\lossce$ is much more unreliable, as higher confidence could indicate greater error rate. By contrast, the model trained with $\loss$ seldom wrongly predicts with a high confidence, highlighting its strong capacity to mitigate overconfidence.

\begin{figure}[t]
    \centering
    \includegraphics[width=0.47\textwidth]{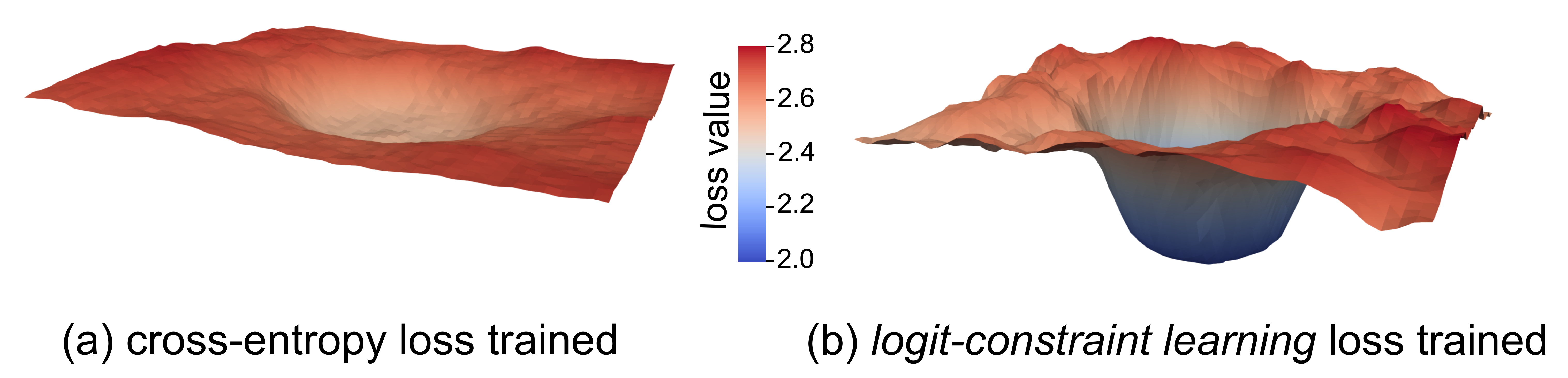}
    \caption{The loss surfaces of models trained with $\lossce$ and $\loss$. Our $\loss$ is more advanced for parameter optimization.}
    \label{fig:landscape}
\end{figure}

\paragraph{Loss Landscape Visualization.} To delve into the optimization potential of \lossModuleName, we plot the loss landscape~\cite{li2018visualizing} of models train with $\lossce$ or $\loss$ in Fig.~\ref{fig:landscape}. The figures are drawn from loss variation with model parameter perturbation, and the statistics is collected on the whole target train set with ground-truth labels. It is clear that the model trained by $\loss$ is able to achieve much lower loss in regions where that trained by $\lossce$ remains confused. Furthermore, our $\loss$ expands the model's potential to pursue superior prediction quality on the target domain, as is illustrated by the blue region featuring minor error.

\paragraph{Influence of Hyperparameters.} We traverse hyperparameters (pseudo threshold $\delta$, scaling factor $\gamma$, and momentum-update ratio $\alpha$) around their optimal values [$\delta^*$, $\gamma^*$, $\alpha^*$]. Results are provided in the Appendix~\ref{sec:parameters}, in which we observe that \method is much less sensitive to its hyperparameters.

\section{Conclusion}

In this paper, we propose \methodblank, an new framework especially designed for better normal-to-adverse adaptation, to explore the possibility of getting rid of reference images. Within this method, contributions are made in both input and output space to enable an improved prediction quality even in poor visibility scenarios. This simple yet effective approach provides the best of both worlds: \boostModuleName~dynamically ameliorates incoming images via certain priors, while \lossModuleName~relieves the pain of overconfidence in the conventional cross-entropy loss for self-training paradigm. Our method can be trained end-to-end in one stage, leading to considerable performance gains on many challenging adverse conditions.


\section*{Acknowledgments}
This paper was supported by National Key R\&D Program of China (No. 2021YFB3301503), and also supported by the National Natural Science Foundation of China under Grant No. U21A20519.

\bibliography{aaai23.bib}

\begin{thebibliography}{65}
\providecommand{\natexlab}[1]{#1}

\bibitem[{Ben-David et~al.(2010)Ben-David, Blitzer, Crammer, Kulesza, Pereira,
  and Vaughan}]{ben2010theory}
Ben-David, S.; Blitzer, J.; Crammer, K.; Kulesza, A.; Pereira, F.; and Vaughan,
  J.~W. 2010.
\newblock A theory of learning from different domains.
\newblock \emph{Mach. Learn.}, 79(1): 151--175.

\bibitem[{Bruggemann et~al.(2022)Bruggemann, Sakaridis, Truong, and
  Van~Gool}]{bruggemann2022refign}
Bruggemann, D.; Sakaridis, C.; Truong, P.; and Van~Gool, L. 2022.
\newblock Refign: Align and Refine for Adaptation of Semantic Segmentation to
  Adverse Conditions.
\newblock \emph{arXiv preprint arXiv:2207.06825}.

\bibitem[{Chen et~al.(2017)Chen, Papandreou, Kokkinos, Murphy, and
  Yuille}]{chen2017deeplab}
Chen, L.-C.; Papandreou, G.; Kokkinos, I.; Murphy, K.; and Yuille, A.~L. 2017.
\newblock Deeplab: Semantic image segmentation with deep convolutional nets,
  atrous convolution, and fully connected crfs.
\newblock \emph{{IEEE} Trans. Pattern Anal. Mach. Intell.}, 40(4): 834--848.

\bibitem[{Chen et~al.(2020)Chen, Fang, Ding, Tsai, and Kuo}]{chen2020jstasr}
Chen, W.-T.; Fang, H.-Y.; Ding, J.-J.; Tsai, C.-C.; and Kuo, S.-Y. 2020.
\newblock JSTASR: Joint size and transparency-aware snow removal algorithm
  based on modified partial convolution and veiling effect removal.
\newblock In \emph{{ECCV}}, 754--770.

\bibitem[{Cordts et~al.(2016)Cordts, Omran, Ramos, Rehfeld, Enzweiler,
  Benenson, Franke, Roth, and Schiele}]{cordts2016cityscapes}
Cordts, M.; Omran, M.; Ramos, S.; Rehfeld, T.; Enzweiler, M.; Benenson, R.;
  Franke, U.; Roth, S.; and Schiele, B. 2016.
\newblock The cityscapes dataset for semantic urban scene understanding.
\newblock In \emph{{CVPR}}, 3213--3223.

\bibitem[{Dai et~al.(2020)Dai, Sakaridis, Hecker, and Gool}]{DaiSHG20}
Dai, D.; Sakaridis, C.; Hecker, S.; and Gool, L.~V. 2020.
\newblock Curriculum Model Adaptation with Synthetic and Real Data for Semantic
  Foggy Scene Understanding.
\newblock \emph{Int. J. Comput. Vis.}, 128(5): 1182--1204.

\bibitem[{Deng et~al.(2009)Deng, Dong, Socher, Li, Li, and
  Fei-Fei}]{deng2009imagenet}
Deng, J.; Dong, W.; Socher, R.; Li, L.-J.; Li, K.; and Fei-Fei, L. 2009.
\newblock Imagenet: A large-scale hierarchical image database.
\newblock In \emph{CVPR}, 248--255.

\bibitem[{Everingham et~al.(2015)Everingham, Eslami, Gool, Williams, Winn, and
  Zisserman}]{everingham2015IoU}
Everingham, M.; Eslami, S. M.~A.; Gool, L.~V.; Williams, C. K.~I.; Winn, J.~M.;
  and Zisserman, A. 2015.
\newblock The Pascal Visual Object Classes Challenge: {A} Retrospective.
\newblock \emph{Int. J. Comput. Vis.}, 111(1): 98--136.

\bibitem[{Ganin et~al.(2016)Ganin, Ustinova, Ajakan, Germain, Larochelle,
  Laviolette, Marchand, and Lempitsky}]{ganin2015dann_jmlr}
Ganin, Y.; Ustinova, E.; Ajakan, H.; Germain, P.; Larochelle, H.; Laviolette,
  F.; Marchand, M.; and Lempitsky, V.~S. 2016.
\newblock Domain-Adversarial Training of Neural Networks.
\newblock \emph{J. Mach. Learn. Res.}, 17(1): 2096--2030.

\bibitem[{Gong et~al.(2021)Gong, Chen, Paudel, Li, Chhatkuli, Li, Dai, and
  Gool}]{GongCPLCLDG21}
Gong, R.; Chen, Y.; Paudel, D.~P.; Li, Y.; Chhatkuli, A.; Li, W.; Dai, D.; and
  Gool, L.~V. 2021.
\newblock Cluster, Split, Fuse, and Update: Meta-Learning for Open Compound
  Domain Adaptive Semantic Segmentation.
\newblock In \emph{CVPR}, 8344--8354.

\bibitem[{He, Sun, and Tang(2009)}]{he2009single}
He, K.; Sun, J.; and Tang, X. 2009.
\newblock Single image haze removal using dark channel prior.
\newblock In \emph{{CVPR}}, 1956--1963.

\bibitem[{He et~al.(2016)He, Zhang, Ren, and Sun}]{he2016deep}
He, K.; Zhang, X.; Ren, S.; and Sun, J. 2016.
\newblock Deep residual learning for image recognition.
\newblock In \emph{CVPR}, 770--778.

\bibitem[{He, Yang, and Qi(2021)}]{he2021re}
He, R.; Yang, J.; and Qi, X. 2021.
\newblock Re-distributing biased pseudo labels for semi-supervised semantic
  segmentation: A baseline investigation.
\newblock In \emph{{ICCV}}, 6930--6940.

\bibitem[{He et~al.(2020)He, Rahimian, Schiele, and
  Fritz}]{he2020segmentations}
He, Y.; Rahimian, S.; Schiele, B.; and Fritz, M. 2020.
\newblock Segmentations-leak: Membership inference attacks and defenses in
  semantic image segmentation.
\newblock In \emph{{ECCV}}, 519--535. Springer.

\bibitem[{Hoffman et~al.(2018)Hoffman, Tzeng, Park, Zhu, Isola, Saenko, Efros,
  and Darrell}]{hoffman2018cycada}
Hoffman, J.; Tzeng, E.; Park, T.; Zhu, J.-Y.; Isola, P.; Saenko, K.; Efros, A.;
  and Darrell, T. 2018.
\newblock Cycada: Cycle-consistent adversarial domain adaptation.
\newblock In \emph{{ICML}}, 1989--1998.

\bibitem[{Hoyer, Dai, and Gool(2022)}]{lukas2021daformer}
Hoyer, L.; Dai, D.; and Gool, L.~V. 2022.
\newblock DAFormer: Improving Network Architectures and Training Strategies for
  Domain-Adaptive Semantic Segmentation.
\newblock In \emph{CVPR}, 9924--9935.

\bibitem[{Hu et~al.(2019)Hu, Fu, Zhu, and Heng}]{hu2019depth}
Hu, X.; Fu, C.-W.; Zhu, L.; and Heng, P.-A. 2019.
\newblock Depth-attentional features for single-image rain removal.
\newblock In \emph{{CVPR}}, 8022--8031.

\bibitem[{Isobe et~al.(2021)Isobe, Jia, Chen, He, Shi, Liu, Lu, and
  Wang}]{isobe2021multi}
Isobe, T.; Jia, X.; Chen, S.; He, J.; Shi, Y.; Liu, J.; Lu, H.; and Wang, S.
  2021.
\newblock Multi-target domain adaptation with collaborative consistency
  learning.
\newblock In \emph{{CVPR}}, 8187--8196.

\bibitem[{Kim and Byun(2020)}]{kim2020learning}
Kim, M.; and Byun, H. 2020.
\newblock Learning texture invariant representation for domain adaptation of
  semantic segmentation.
\newblock In \emph{{CVPR}}, 12975--12984.

\bibitem[{Kingma and Ba(2015)}]{KingmaB14}
Kingma, D.~P.; and Ba, J. 2015.
\newblock Adam: {A} Method for Stochastic Optimization.
\newblock In \emph{ICLR}.

\bibitem[{Land(1977)}]{land1977retinex}
Land, E.~H. 1977.
\newblock The retinex theory of color vision.
\newblock \emph{Scientific american}, 237(6): 108--129.

\bibitem[{Lee et~al.(2022)Lee, Choi, Kim, Choi, and Im}]{lee2022adas}
Lee, S.; Choi, W.; Kim, C.; Choi, M.; and Im, S. 2022.
\newblock ADAS: A Direct Adaptation Strategy for Multi-Target Domain Adaptive
  Semantic Segmentation.
\newblock In \emph{{CVPR}}, 19196--19206.

\bibitem[{Li et~al.(2018)Li, Xu, Taylor, Studer, and
  Goldstein}]{li2018visualizing}
Li, H.; Xu, Z.; Taylor, G.; Studer, C.; and Goldstein, T. 2018.
\newblock Visualizing the loss landscape of neural nets.
\newblock In \emph{NeurIPS}, 6391--6401.

\bibitem[{Li, Cheong, and Tan(2019)}]{li2019heavy}
Li, R.; Cheong, L.-F.; and Tan, R.~T. 2019.
\newblock Heavy rain image restoration: Integrating physics model and
  conditional adversarial learning.
\newblock In \emph{{CVPR}}, 1633--1642.

\bibitem[{Li et~al.(2022)Li, Xie, Lin, Liu, Huang, and Wang}]{GDCAN}
Li, S.; Xie, B.; Lin, Q.; Liu, C.~H.; Huang, G.; and Wang, G. 2022.
\newblock Generalized Domain Conditioned Adaptation Network.
\newblock \emph{IEEE Trans. Pattern Anal. Mach. Intell.}, 44(8): 4093--4109.

\bibitem[{Li, Yuan, and Vasconcelos(2019)}]{li2019bidirectional}
Li, Y.; Yuan, L.; and Vasconcelos, N. 2019.
\newblock Bidirectional learning for domain adaptation of semantic
  segmentation.
\newblock In \emph{{CVPR}}, 6936--6945.

\bibitem[{Liu et~al.(2022)Liu, Ren, Yu, Guo, Zhu, and Zhang}]{liu2022image}
Liu, W.; Ren, G.; Yu, R.; Guo, S.; Zhu, J.; and Zhang, L. 2022.
\newblock Image-Adaptive {YOLO} for Object Detection in Adverse Weather
  Conditions.
\newblock In \emph{{AAAI}}, 1792--1800.

\bibitem[{Liu et~al.(2020)Liu, Miao, Pan, Zhan, Lin, Yu, and
  Gong}]{liu2020ocda}
Liu, Z.; Miao, Z.; Pan, X.; Zhan, X.; Lin, D.; Yu, S.~X.; and Gong, B. 2020.
\newblock Open Compound Domain Adaptation.
\newblock In \emph{CVPR}, 12403--12412.

\bibitem[{Long et~al.(2018)Long, Cao, Wang, and Jordan}]{long2018conditional}
Long, M.; Cao, Z.; Wang, J.; and Jordan, M.~I. 2018.
\newblock Conditional Adversarial Domain Adaptation.
\newblock In \emph{NeurIPS}, 1647--1657.

\bibitem[{Loshchilov and Hutter(2019)}]{loshchilov2017decoupled}
Loshchilov, I.; and Hutter, F. 2019.
\newblock Decoupled weight decay regularization.
\newblock In \emph{{ICLR}}. OpenReview.net.

\bibitem[{Luo et~al.(2019)Luo, Zheng, Guan, Yu, and Yang}]{luo2019taking}
Luo, Y.; Zheng, L.; Guan, T.; Yu, J.; and Yang, Y. 2019.
\newblock Taking a closer look at domain shift: Category-level adversaries for
  semantics consistent domain adaptation.
\newblock In \emph{{CVPR}}, 2507--2516.

\bibitem[{Ma et~al.(2022)Ma, Wang, Zhan, Zheng, Wang, Dai, and
  Lin}]{ma2022both}
Ma, X.; Wang, Z.; Zhan, Y.; Zheng, Y.; Wang, Z.; Dai, D.; and Lin, C.-W. 2022.
\newblock Both style and fog matter: Cumulative domain adaptation for semantic
  foggy scene understanding.
\newblock In \emph{CVPR}, 18922--18931.

\bibitem[{McCartney(1976)}]{mccartney1976optics}
McCartney, E.~J. 1976.
\newblock Optics of the atmosphere: scattering by molecules and particles.
\newblock \emph{New York}.

\bibitem[{Narasimhan and Nayar(2003)}]{narasimhan2003contrast}
Narasimhan, S.~G.; and Nayar, S.~K. 2003.
\newblock Contrast restoration of weather degraded images.
\newblock \emph{{IEEE} Trans. Pattern Anal. Mach. Intell.}, 25(6): 713--724.

\bibitem[{Nayar and Narasimhan(1999)}]{nayar1999vision}
Nayar, S.~K.; and Narasimhan, S.~G. 1999.
\newblock Vision in bad weather.
\newblock In \emph{{ICCV}}, volume~2, 820--827.

\bibitem[{Neuhold et~al.(2017)Neuhold, Ollmann, Rota~Bulo, and
  Kontschieder}]{neuhold2017mapillary}
Neuhold, G.; Ollmann, T.; Rota~Bulo, S.; and Kontschieder, P. 2017.
\newblock The mapillary vistas dataset for semantic understanding of street
  scenes.
\newblock In \emph{{ICCV}}, 4990--4999.

\bibitem[{Olsson et~al.(2021)Olsson, Tranheden, Pinto, and
  Svensson}]{olsson2021classmix}
Olsson, V.; Tranheden, W.; Pinto, J.; and Svensson, L. 2021.
\newblock Classmix: Segmentation-based data augmentation for semi-supervised
  learning.
\newblock In \emph{{WACV}}, 1369--1378.

\bibitem[{Park et~al.(2020)Park, Woo, Shin, and Kweon}]{ParkWSK20}
Park, K.; Woo, S.; Shin, I.; and Kweon, I.~S. 2020.
\newblock Discover, Hallucinate, and Adapt: Open Compound Domain Adaptation for
  Semantic Segmentation.
\newblock In \emph{NeurIPS}.

\bibitem[{Paszke et~al.(2019)Paszke, Gross, Massa, Lerer, Bradbury, Chanan,
  Killeen, Lin, Gimelshein, Antiga, Desmaison, K{\"{o}}pf, Yang, DeVito,
  Raison, Tejani, Chilamkurthy, Steiner, Fang, Bai, and
  Chintala}]{paszke2019pytorch}
Paszke, A.; Gross, S.; Massa, F.; Lerer, A.; Bradbury, J.; Chanan, G.; Killeen,
  T.; Lin, Z.; Gimelshein, N.; Antiga, L.; Desmaison, A.; K{\"{o}}pf, A.; Yang,
  E.; DeVito, Z.; Raison, M.; Tejani, A.; Chilamkurthy, S.; Steiner, B.; Fang,
  L.; Bai, J.; and Chintala, S. 2019.
\newblock PyTorch: An Imperative Style, High-Performance Deep Learning Library.
\newblock In \emph{NeurIPS}, 8024--8035.

\bibitem[{Sakaridis, Dai, and Gool(2022)}]{sakaridis2020map}
Sakaridis, C.; Dai, D.; and Gool, L.~V. 2022.
\newblock Map-Guided Curriculum Domain Adaptation and Uncertainty-Aware
  Evaluation for Semantic Nighttime Image Segmentation.
\newblock \emph{{IEEE} Trans. Pattern Anal. Mach. Intell.}, 44(6): 3139--3153.

\bibitem[{Sakaridis et~al.(2018)Sakaridis, Dai, Hecker, and
  Van~Gool}]{sakaridis2018model}
Sakaridis, C.; Dai, D.; Hecker, S.; and Van~Gool, L. 2018.
\newblock Model adaptation with synthetic and real data for semantic dense
  foggy scene understanding.
\newblock In \emph{{ECCV}}, 687--704.

\bibitem[{Sakaridis, Dai, and Van~Gool(2018)}]{sakaridis2018semantic}
Sakaridis, C.; Dai, D.; and Van~Gool, L. 2018.
\newblock Semantic foggy scene understanding with synthetic data.
\newblock \emph{Int. J. Comput. Vis.}, 126(9): 973--992.

\bibitem[{Sakaridis, Dai, and Van~Gool(2021)}]{sakaridis2021acdc}
Sakaridis, C.; Dai, D.; and Van~Gool, L. 2021.
\newblock ACDC: The adverse conditions dataset with correspondences for
  semantic driving scene understanding.
\newblock In \emph{{ICCV}}, 10765--10775.

\bibitem[{Saporta et~al.(2021)Saporta, Vu, Cord, and
  P{\'e}rez}]{saporta2021multi}
Saporta, A.; Vu, T.-H.; Cord, M.; and P{\'e}rez, P. 2021.
\newblock Multi-target adversarial frameworks for domain adaptation in semantic
  segmentation.
\newblock In \emph{{ICCV}}, 9072--9081.

\bibitem[{Tranheden et~al.(2021)Tranheden, Olsson, Pinto, and
  Svensson}]{tranheden2021dacs}
Tranheden, W.; Olsson, V.; Pinto, J.; and Svensson, L. 2021.
\newblock DACS: Domain adaptation via cross-domain mixed sampling.
\newblock In \emph{{WACV}}, 1379--1389.

\bibitem[{Tsai et~al.(2018)Tsai, Hung, Schulter, Sohn, Yang, and
  Chandraker}]{tsai2018learning}
Tsai, Y.-H.; Hung, W.-C.; Schulter, S.; Sohn, K.; Yang, M.-H.; and Chandraker,
  M. 2018.
\newblock Learning to adapt structured output space for semantic segmentation.
\newblock In \emph{{CVPR}}, 7472--7481.

\bibitem[{Tzeng et~al.(2017)Tzeng, Hoffman, Saenko, and
  Darrell}]{tzeng2017adversarial}
Tzeng, E.; Hoffman, J.; Saenko, K.; and Darrell, T. 2017.
\newblock Adversarial discriminative domain adaptation.
\newblock In \emph{CVPR}, 7167--7176.

\bibitem[{Valanarasu, Yasarla, and Patel(2022)}]{valanarasu2022transweather}
Valanarasu, J. M.~J.; Yasarla, R.; and Patel, V.~M. 2022.
\newblock Transweather: Transformer-based restoration of images degraded by
  adverse weather conditions.
\newblock In \emph{{CVPR}}, 2353--2363.

\bibitem[{Varma et~al.(2019)Varma, Subramanian, Namboodiri, Chandraker, and
  Jawahar}]{varma2019idd}
Varma, G.; Subramanian, A.; Namboodiri, A.; Chandraker, M.; and Jawahar, C.
  2019.
\newblock IDD: A dataset for exploring problems of autonomous navigation in
  unconstrained environments.
\newblock In \emph{{WACV}}, 1743--1751.

\bibitem[{Vu et~al.(2019)Vu, Jain, Bucher, Cord, and P{\'e}rez}]{vu2019advent}
Vu, T.-H.; Jain, H.; Bucher, M.; Cord, M.; and P{\'e}rez, P. 2019.
\newblock Advent: Adversarial entropy minimization for domain adaptation in
  semantic segmentation.
\newblock In \emph{{CVPR}}, 2517--2526.

\bibitem[{Wang and Deng(2018)}]{WangD18}
Wang, M.; and Deng, W. 2018.
\newblock Deep visual domain adaptation: {A} survey.
\newblock \emph{Neurocomputing}, 312: 135--153.

\bibitem[{Wang et~al.(2021{\natexlab{a}})Wang, Dai, Hoyer, Van~Gool, and
  Fink}]{wang2021domain}
Wang, Q.; Dai, D.; Hoyer, L.; Van~Gool, L.; and Fink, O. 2021{\natexlab{a}}.
\newblock Domain adaptive semantic segmentation with self-supervised depth
  estimation.
\newblock In \emph{{ICCV}}, 8515--8525.

\bibitem[{Wang et~al.(2021{\natexlab{b}})Wang, Cao, Zhang, He, Zha, Wen, and
  Tao}]{Wang2021SFA}
Wang, W.; Cao, Y.; Zhang, J.; He, F.; Zha, Z.; Wen, Y.; and Tao, D.
  2021{\natexlab{b}}.
\newblock Exploring Sequence Feature Alignment for Domain Adaptive Detection
  Transformers.
\newblock In \emph{{ACM MM}}, 1730--1738.

\bibitem[{Wei et~al.(2022)Wei, Xie, Cheng, Feng, An, and
  Li}]{wei2022mitigating}
Wei, H.; Xie, R.; Cheng, H.; Feng, L.; An, B.; and Li, Y. 2022.
\newblock Mitigating Neural Network Overconfidence with Logit Normalization.
\newblock \emph{arXiv preprint arXiv:2205.09310}.

\bibitem[{Wu et~al.(2021)Wu, Wu, Ju, and Wang}]{wu2021one}
Wu, X.; Wu, Z.; Ju, L.; and Wang, S. 2021.
\newblock A One-Stage Domain Adaptation Network with Image Alignment for
  Unsupervised Nighttime Semantic Segmentation.
\newblock \emph{{IEEE} Trans. Pattern Anal. Mach. Intell.}, 1--1.

\bibitem[{Wulfmeier, Bewley, and Posner(2018)}]{wulfmeier2018incremental}
Wulfmeier, M.; Bewley, A.; and Posner, I. 2018.
\newblock Incremental adversarial domain adaptation for continually changing
  environments.
\newblock In \emph{{ICRA}}, 4489--4495.

\bibitem[{Xie et~al.(2022)Xie, Li, Li, Liu, Huang, and Wang}]{xie2022sepico}
Xie, B.; Li, S.; Li, M.; Liu, C.~H.; Huang, G.; and Wang, G. 2022.
\newblock SePiCo: Semantic-Guided Pixel Contrast for Domain Adaptive Semantic
  Segmentation.
\newblock \emph{arXiv preprint arXiv:2204.08808}.

\bibitem[{Xie et~al.(2021)Xie, Wang, Yu, Anandkumar, Alvarez, and
  Luo}]{xie2021segformer}
Xie, E.; Wang, W.; Yu, Z.; Anandkumar, A.; Alvarez, J.~M.; and Luo, P. 2021.
\newblock SegFormer: Simple and efficient design for semantic segmentation with
  transformers.
\newblock \emph{NeurIPS}.

\bibitem[{Yang and Soatto(2020)}]{yang2020fda}
Yang, Y.; and Soatto, S. 2020.
\newblock {FDA:} Fourier Domain Adaptation for Semantic Segmentation.
\newblock In \emph{{CVPR}}, 4085--4095.

\bibitem[{Zhang et~al.(2021{\natexlab{a}})Zhang, Zhang, Zhang, Chen, Wang, and
  Wen}]{zhang2021prototypical}
Zhang, P.; Zhang, B.; Zhang, T.; Chen, D.; Wang, Y.; and Wen, F.
  2021{\natexlab{a}}.
\newblock Prototypical pseudo label denoising and target structure learning for
  domain adaptive semantic segmentation.
\newblock In \emph{{CVPR}}, 12414--12424.

\bibitem[{Zhang et~al.(2012)Zhang, Shen, Luo, Zhang, and
  Song}]{zhang2012enhancement}
Zhang, X.; Shen, P.; Luo, L.; Zhang, L.; and Song, J. 2012.
\newblock Enhancement and noise reduction of very low light level images.
\newblock In \emph{{ICPR}}, 2034--2037.

\bibitem[{Zhang et~al.(2021{\natexlab{b}})Zhang, Carballo, Yang, and
  Takeda}]{zhang2021autonomous}
Zhang, Y.; Carballo, A.; Yang, H.; and Takeda, K. 2021{\natexlab{b}}.
\newblock Autonomous Driving in Adverse Weather Conditions: A Survey.
\newblock \emph{arXiv preprint arXiv:2112.08936}.

\bibitem[{Zhu et~al.(2021)Zhu, Su, Lu, Li, Wang, and Dai}]{DeformableDETR}
Zhu, X.; Su, W.; Lu, L.; Li, B.; Wang, X.; and Dai, J. 2021.
\newblock Deformable {DETR:} Deformable Transformers for End-to-End Object
  Detection.
\newblock In \emph{ICLR}.

\bibitem[{Zou et~al.(2018)Zou, Yu, Kumar, and Wang}]{zou2018unsupervised}
Zou, Y.; Yu, Z.; Kumar, B.; and Wang, J. 2018.
\newblock Unsupervised domain adaptation for semantic segmentation via
  class-balanced self-training.
\newblock In \emph{{ECCV}}, 289--305.

\bibitem[{Zou et~al.(2019)Zou, Yu, Liu, Kumar, and Wang}]{zou2019confidence}
Zou, Y.; Yu, Z.; Liu, X.; Kumar, B.; and Wang, J. 2019.
\newblock Confidence regularized self-training.
\newblock In \emph{{ICCV}}, 5982--5991.

\end{thebibliography}

\clearpage

\appendix
\section*{Appendix}

\paragraph{Contents}
\begin{itemize}
    \item Experiment Steup~\ref{sec:experiemnt_steup}
    \begin{itemize}
      \item Dataset Details~\ref{sec:dataset}
      \item Implementation Details~\ref{sec:implementation}
    \end{itemize}
    \item \method Algorithm~\ref{sec:alg}
    \item Mathematical Derivations~\ref{sec:proof}
    \item Additional Results~\ref{sec:additional_results}
    \begin{itemize}
      \item More Quantitative results~\ref{sec:more_quantitative}
      \item Results on Hyperparameters~\ref{sec:parameters}
      \item More Qualitative Results~\ref{sec:more_qualitative}
      \item Failure Cases~\ref{sec:failure}
    \end{itemize}
\end{itemize}

\section{Experiment Setup}
\label{sec:experiemnt_steup}

\subsection{Dataset Details}
\label{sec:dataset}

\noindent\textbf{Cityscapes}~\cite{cordts2016cityscapes} is a real-world urban scene dataset consisting of daytime images from 50 cities. The shooting season could vary greatly, but the taken images are generally in good or medium weather conditions. Therefore, we view this dataset as one with appropriate visibility, namely in `clear' condition. As for experiments, we use the finely annotated subset, which contains 2,975 images for training, 500 for validation, and 1,525 for test purpose, with a resolution of 2048$\times$1024. The pixel-wise annotation consists of 19 semantic classes, as per standard practice. We use the training images with their annotations as the source domain for both tasks.

\noindent\textbf{ACDC}~\cite{sakaridis2021acdc} is another dataset consisting of real images recorded in Switzerland, primarily in urban areas. The ACDC dataset shares the same 19 class with Cityscapes, but is explicitly divided into four adverse conditions, namely fog, night, rain, and snow. For each adverse split, 400 training images, 100 validation images (except for night, which has 106 for validation), and 500 test images are provided with 1920$\times$1080 resolution. Although all these images are well-annotated, we only use the 1,600 training images without their labels as the target domain. We tune the hyperparameters on ACDC's validation set, and the final results are reported on its test set. The test results are obtained through the evaluation server\footnote{https://acdc.vision.ee.ethz.ch/submit}, as the annotations of test images are withheld.

\noindent\textbf{FoggyCityscapes}~\cite{sakaridis2018semantic} is a synthetic foggy dataset deriving from the Cityscapes dataset, and as such the annotations are automatically inherited. Foggy images with three different visibility ranges are generated for every single image. In our experiments, all these three varieties of training data are combined to form a target domain with 8,925 unlabeled images. Also, the test results are based on the expanded validation set of 1,500 annotated images.

\noindent\textbf{RainCityscapes}~\cite{hu2019depth} is another synthetic rain dataset modified from a subset of Cityscapes dataset. A certain clear image from Cityscapes is mapped to 36 different rainy images to simulate diverse degrees of rain and fog. Therefore, a total of 9,432 training images and 1,188 validation images are included. We use this dataset together with FoggyCityscapes as the target domain, and test results on the validation set.

\noindent\textbf{IDD}~\cite{varma2019idd}, short for India Driving Dataset, is a driving dataset featuring unstructured environments collected on Indian roads. The images are mostly at a resolution of 1920$\times$1080, but other resolutions exist, including 1280$\times$720. As to the dataset structure, a total of 10,003 images are split into a training set of 6,993 images, a validation set of 981 images, and a testing set of 2,029 images. For semantic segmentation task, an annotation compatible with Cityscapes is provided, and we only use the shared 19 classes for our experiments. Moreover, for the 7 super classes setting, we perform a mapping following MTKT~\cite{saporta2021multi}. For the multi-target domain adaptive semantic segmentation task, we only use the training images without their labels, and evaluate on the validation set. 

\noindent\textbf{Mapillary Vista}~\cite{neuhold2017mapillary} is a large-scale street-level image dataset captured worldwide. The images are of high resolution, generally above 1920$\times$1080 and mainly around 4K resolution. 25,000 images make up the full dataset, and are divided into 18,000 for training, 2,000 for validation, and 5,000 for testing. We adopt the v1.2 version for a fair comparison with existing methods, whose annotation consists of 66 object categories. To enable the adaptation, we map them to the 19 classes in Cityscapes according to~\cite{he2020segmentations}, and the mapping to 7 classes is in line with IDD. Only training images along with validation ones are used in our multi-target semantic segmentation task, just like how we use IDD.

\subsection{Implementation Details}
\label{sec:implementation}
All experiments are conducted on a Tesla V100 GPU with PyTorch~\cite{paszke2019pytorch}.

\paragraph{Task 1: Cityscapes $\to$ ACDC semantic segmentation task.}
We thank the mmsegmentation\footnote{https://github.com/open-mmlab/mmsegmentation} toolbox, as our implementation regarding semantic segmentation is heavily reliant on it.
In this task, we adopt ResNet-101~\cite{he2016deep} + DeepLab-v2~\cite{chen2017deeplab} as well as MiT-B5~\cite{xie2021segformer} + DAFormer~\cite{lukas2021daformer} as the architecture. Both ResNet-101 and MiT-B5 backbones are pretrained on ImageNet~\cite{deng2009imagenet}. For both architectures, AdamW~\cite{loshchilov2017decoupled} with betas $(0.9, 0.999)$ is used as the optimimizer with a 0.01 weight decay, and the network is trained for 40k iterations. The batch size is 4, namely 2 source images with 2 target images. The initial learning rate is set to $6\times 10^{-5}$ for the encoder, and $6\times 10^{-4}$ for the decoder. Learning rate warmup policy and rare class sampling are borrowed from~\cite{lukas2021daformer} for better transferability. We adopt the poly schedule with a power of 1.0 for learning rate update.

During training, data augmentations including Resize, Random Crop, Random Flip, Gaussian blur, and Color Jitter are applied, and the image size for training is 640$\times$640. Hyperparameters are set to $\delta=0.9, \gamma=4.0, \alpha=0.999$, respectively. For testing, the images are first resized to 1280$\times$720 as the input. Only the student model is necessary for the test stage, as the visibility gap is closed by the simultaneous training on both the original target image and the boosted one, while the constraint on logit would not influence the outcome of a simple $argmax$ inference strategy.

We compare with existing domain adaptive semantic segmentation methods, including ResNet-101 based methods DeepLab-v2~\cite{chen2017deeplab}, AdaptSegNet~\cite{tsai2018learning}, ADVENT~\cite{vu2019advent}, BDL~\cite{li2019bidirectional}, CLAN~\cite{luo2019taking}, CRST~\cite{zou2019confidence}, FDA~\cite{yang2020fda}, DACS~\cite{tranheden2021dacs}, and MiT-B5 based methods SegFormer~\cite{xie2021segformer}, DAFormer~\cite{lukas2021daformer}. Among them, DeepLab-v2 and SegFormer results are attained by testing their Source-only models on the target domain, while others are all domain adaptation methods.

\paragraph{Task 2: Cityscapes $\to$ FoggyCityscapes + RainCityscapes semantic segmentation task.}
The implementation for this task is basically the same with that of Task 1, except for the testing image size is set to 1280$\times$640. We would like to give special clarification that, although images in the target domain of this task come from two distinct datasets, we use the direct combination of the dataset and treat them as one single dataset for sampling. The evaluation is also made on the combined dataset.

We implement several domain adaptive semantic segmentation methods according to source code, including DeepLab-v2~\cite{chen2017deeplab}, FDA~\cite{yang2020fda}\footnote{https://github.com/YanchaoYang/FDA}, DACS~\cite{tranheden2021dacs}\footnote{https://github.com/vikolss/DACS}, SegFormer~\cite{xie2021segformer}\footnote{https://github.com/NVlabs/SegFormer}, DAFormer~\cite{lukas2021daformer}\footnote{https://github.com/lhoyer/DAFormer}.

\paragraph{Task 3: Cityscapes $\to$ FoggyCityscapes + RainCityscapes object detection task.} 
To showcase the flexibility of \methodblank, we also focus on object detection task under adverse conditions.
We adopt \boostModuleName~and \lossModuleName~as incremental modules to a state-of-the-art domain adaptive object detection method, SFA~\cite{Wang2021SFA} without any other modifications. Specifically,  images from the target domain is first boosted with the designed \boostModuleName, which performs coarse alignment between normal and adverse conditions. Then, the \lossModuleName~is integrated with class prediction.

As for Task 2, we use all images in FoggyCityscapes and RainCityscapes and merge both datasets into a combined target domain. The evaluation is conducted on validation sets of the two datasets. We use the same setting as SFA~\cite{Wang2021SFA}, which is build on DeformableDETR~\cite{DeformableDETR}. ImageNet~\cite{deng2009imagenet} pre-trained ResNet-50~\cite{he2016deep} is adopted as the backbone for all comparison approaches. Follwing SFA~\cite{Wang2021SFA}, we train the network using Adam optimimizer~\cite{KingmaB14} for 50 epochs. The batch size is 4 and the intial learning rate is set as $2\times 10^{-4}$. Learning rate is decayed by 0.1 after 40 epochs. 

We implement two baseline methods (DeformableDETR and SFA) and our \method based on source code of SFA\footnote{https://github.com/encounter1997/SFA/tree/main}.

\paragraph{Task 4: Cityscapes $\to$ IDD + Mapillary semantic segmentation task.} Despite the fact that target domains in multi-target domain adaptations are frequently sampled separately, we continue to employ our own pipeline and merge IDD and Mapillary into a combined dataset, which is much similar to the practice of ADAS~\cite{lee2022adas}. Therefore, the implementation details of Task 1 are still applicable to this Task. The testing image size is set to 1280$\times$720 for IDD, and 1280$\times$640 for Mapillary.

We compare with existing multi-target domain adaptive semantic segmentation methods, including MTKT~\cite{saporta2021multi}, ADAS~\cite{lee2022adas}, CCL~\cite{isobe2021multi}, on both 19 classes and 7 super classes.

\begin{algorithm}[t]
  \begin{algorithmic}[1] 
  \caption{\method algorithm}
  \label{alg:algorithm}
  \STATE \textbf{Input:} Labeled source domain $\mathcal{S}$, unlabeled target domain $\mathcal{T}$, maximum iteration $R$, ImageNet pretrained student model $\Phi$, momentum teacher model $\Phi'$. \\
  \STATE \textbf{Output:} Final model parameters $\Phi$.
  \STATE Initiate teacher model $\Phi'$ with $\Phi$.
  \FOR{$m=1$ to $R$}
  \STATE Update teacher model $\Phi'$ with $\Phi$ using a momentum scheme: $\Phi' = \alpha\Phi' + (1-\alpha)\Phi$.
  \STATE Sample training data $I_s, Y_s \in \mathcal{S}, I_t \in \mathcal{T}$.
  \STATE Feed target data $I_t$ into $\boostModuleName$ to get enhanced data $\boostimage$.
  \STATE Predict target pseudo label $\targetpseudo$ from $P_b$, which is obtained from $\boostimage$ through teacher model $\Phi'$.
  \STATE Mix up source-target image pairs, i.e., $\mixtargetimage$ and $\mixboostimage$, and mixed label $\mixlabel$ for self-training.
  \STATE Compute logits $\sourcelogit, \mixtargetlogit, \mixboostlogit$ from $\sourceimage, \mixtargetimage, \mixboostimage$ through student model $\Phi$, respectively.
  \STATE Normalize logits through \lossModuleName~to get $\sourcelogitnorm, \mixtargetlogitnorm, \mixboostlogitnorm$ and further employ softmax to get predictions $\sourceprob,\mixtargetprob, \mixboostprob$.
  \STATE Train $\Phi$ via Eq.~\eqref{eq:overall_loss}.
  \ENDFOR
  \end{algorithmic}
\end{algorithm}

\section{\method Algorithm}
\label{sec:alg}
In this section, we will give a detailed description of the training process. Essentially, a pair of teacher-student models are adopted for self-training. The teacher network is initialized by the student model, and is updated by the student model in the beginning of every iteration using a momentum scheme. For a labeled source domain $\mathcal{S}$ and an unlabeled target domain $\mathcal{T}$, we first randomly sample a source image $I_s$ with its ground-truth label $Y_s$ and a target image $I_t$ without label. We then pass $I_t$ through \boostModuleName~to acquire a boosted target image $I_b$. With $I_b$ in hand, we utilize the teacher model $\Phi'$ to calculate its probability map $P_b$, from which the shared pseudo label $\hat{Y}_b$ is predicted. Subsequently, we perform a ClassMix to mix up the source-target image pairs, where $I_t$ and $I_s$ are mixed to get $\mixtargetimage$, while $I_b$ and $I_s$ are mixed to get $\mixboostimage$. Next, we use the student model $\Phi$ to compute logits $Z_s, \mixtargetlogit, \mixboostlogit$ from $I_s, \mixtargetimage, \mixboostimage$, respectively. These logits are then normalized to $\sourcelogitnorm, \mixtargetlogitnorm, \mixboostlogitnorm$ along the channel dimension according to \lossModuleName, and corresponding predictions are noted as $\sourceprob,\mixtargetprob, \mixboostprob$. Ultimately, the student model is optimized using the loss function in Eq.~\eqref{eq:overall_loss}.
We summarize the whole training procedure in Alg.~\ref{alg:algorithm}. 

\section{Mathematical Derivations}
\label{sec:proof}

\allowdisplaybreaks

\newcommand{\znorm}{{\Vert z\Vert}}
\newcommand{\zdiv}[1]{{z_{#1}/\znorm}}
\newcommand{\zdivf}[1]{{\frac{z_{#1}}{\znorm}}}
\newcommand{\ksum}[1][k]{{\sum_{{#1}=1}^K\ }}
\newcommand{\softmax}[3][k]{{\frac{e^{#2}}{\ksum[#1] e^{#3}}}}
\newcommand{\partialf}[2]{{}\frac{\partial {#1}}{\partial {#2}}}
\newcommand{\p}[2][*]{{p_{#2}^{#1}}}

\paragraph{Derivation of the Gradient of $\lossce$.}
For any pixel $x$ assigned with class $c$ in an image, we use $y$ to denote its one-hot annotation, namely $y_c=1$, and $y_k=0$ holds for any $k\neq c$. Let $z$ be the logit output of the model on input $x$, and $z_i$ is the $i^{th}$ element of it. Per-class prediction $p$ is calculated by applying softmax to the logit $z$. 

Formally, the vanilla cross-entropy loss is formulated as:
\begin{small}
  \begin{equation}
    \begin{aligned}
      \lossce = - \ksum y_k \log(\p[]{k}),~\text{where}~~\p[]{i} = \softmax{z_i}{z_k}\,, \\
    \end{aligned}
  \end{equation}
\end{small}%
and its gradient to a logit element $z_j$ is presented as:
\begin{small}
  \begin{equation}
      \begin{aligned}
      \frac{\partial \mathcal{L}_{ce}}{\partial z_j} = p_j - y_j\,.
      \end{aligned}
  \end{equation}
\end{small}%
As first, we give the derivation of the gradient of $\lossce$ to $z_j$, the $j^{th}$ element of logit, through the chain rule:
\begin{small}
  \begin{equation}
    \begin{aligned}
      \partialf{\lossce}{z_j} &= - \ksum y_k \cdot \partialf{\log(\p[]{k})}{z_j} \\
      &= - \ksum y_k \cdot \partialf{\log(\p[]{k})}{\p[]{k}} \cdot \partialf{\p[]{k}}{z_j} \\
      &= - \ksum y_k \cdot \frac{1}{\p[]{k}} \cdot \partialf{\p[]{k}}{z_j} \\
      &= - y_j \cdot \frac{1}{\p[]{j}} \cdot \partialf{\p[]{j}}{z_j} - \sum_{k\neq j} y_k \cdot \frac{1}{\p[]{k}} \cdot \partialf{\p[]{k}}{z_j} \,. \\
      \label{eq:ce_grad_supp}      
    \end{aligned}
  \end{equation}
\end{small}%
The required gradient of prediction $p_i$ to $z_j$ is calculated as:
\begin{small}
  \begin{equation}
    \begin{aligned}
      \partialf{p_i}{z_j} &= \partialf{}{z_j} \left( \softmax{z_i}{z_k} \right) \\
      &= \frac{ \partialf{}{z_j}(e^{z_i}) \ksum e^{z_k} - e^{z_i} \partialf{}{z_j}\left(\ksum e^{z_k}\right) }{\left( \ksum e^{z_k} \right)^2}  \,,\\
    \end{aligned}
  \end{equation}
\end{small}%
Then, the gradient of prediction to logit is yielded as
\begin{small}
  \begin{equation}
    \partialf{\p[]{i}}{z_j} = \left\{
    \begin{aligned}
      & \p[]{j} (1 - \p[]{j}), i = j \\
      & \p[]{i} ( - \p[]{j}), i\neq j \,.
    \end{aligned}
    \right. 
    \label{eq:ce_pred_grad}
  \end{equation}
\end{small}%
Finally, substitute Eq.~\eqref{eq:ce_pred_grad} back into Eq.~\eqref{eq:ce_grad_supp}, and we have
\begin{small}
  \begin{equation}
    \begin{aligned}
      \partialf{\lossce}{z_j} &= - y_j \cdot \frac{1}{p_j} \cdot \p[]{j} (1 - \p[]{j}) - \sum_{k\neq j} y_k \cdot \frac{1}{\p[]{k}} \cdot \p[]{k}(- \p[]{j}) \\
      &= - y_j + \p[]{j} \left(\ksum y_k \right) \\
      &= \p[]{j} - y_j  \,.
    \end{aligned}
  \end{equation}
\end{small}

\paragraph{Derivation of the Gradient of $\loss$.}
The derivation for $\loss$ is similar to that of $\lossce$, but a bit more complex. We use the same notations as those in the derivation of $\lossce$, but with two exceptions: we now introduce the $\ell_2$-norm of the logit, namely $\Vert z\Vert$, and use $\p{i}$ to denote the new prediction for a distinction. Mathematically, the proposed \lossModuleName~loss is formulated as follows:
\begin{small}
  \begin{equation}
    \begin{aligned}
        \loss = - \ksum y_k\log(\p{k}),~\text{where}~~\p{i} = \softmax{\zdiv{i}}{\zdiv{k}}\,, \\
    \end{aligned}
  \end{equation}
\end{small}%
and its corresponding gradient to the $j^{th}$ element of $z$ is given as:
\begin{small}
  \begin{equation}
      \begin{aligned}
      \frac{\partial \loss}{\partial z_j} = \frac{1}{\Vert z \Vert}\left((p_j^* - y_j) - \sum\limits_{k=1}^K \frac{z_j z_k}{{\Vert z \Vert}^2}(p_k^* - y_k)\right)\,.
      \end{aligned}
  \end{equation}
\end{small}%
Similarly, the gradient of $\loss$ to $z_j$ is first derivated by:
\begin{small}
  \begin{equation}
    \begin{aligned}
        \partialf{\loss}{z_j} &= - \ksum y_k \cdot \partialf{\log(\p{k})}{z_j} \\
        &= -\ksum y_k \cdot \partialf{\log(\p{k})}{\p{k}} \cdot \partialf{\p{k}}{z_j} \\
        &= -\ksum y_k \cdot \frac{1}{\p{k}} \cdot \partialf{\p{k}}{z_j} \\
        &= - y_j \cdot \frac{1}{\p{j}} \cdot \partialf{\p{j}}{z_j} - \sum_{k\neq j} y_k \cdot \frac{1}{\p{k}} \cdot \partialf{\p{k}}{z_j} \,.\\
    \end{aligned}
  \end{equation}
\end{small}%
Next, the gradient of prediction to logit is then given as:
\begin{small}
  \begin{equation}
    \begin{aligned}
      \partialf{p_i}{z_j} &= \partialf{}{z_j} \left( \softmax{\zdiv{i}}{\zdiv{k}} \right) \\
      &= \frac{ \partialf{}{z_j}\left(e^{\zdiv{i}}\right)\ksum e^{\zdiv{k}} - e^{\zdiv{i}}\partialf{}{z_j} \left(\ksum e^{\zdiv{k}}\right)  }{\left(\ksum e^{\zdiv{k}}\right)^2} \,.\\
      \label{eq:lc_grad_supp}
    \end{aligned}
  \end{equation}
\end{small}%
To obtain the outcomes of partial derivatives present, we deduce them separately.
On the one hand, the derivation of the former gradient is as follows:
\begin{small}
  \begin{equation}
    \begin{aligned}
      &\partialf{}{z_j} \left( e^{\zdiv{i}} \right) = e^{\zdiv{i}} \cdot \partialf{}{z_j} \left( \zdivf{i} \right) = e^{\zdiv{i}} \cdot \frac{\partialf{z_i}{z_j} \znorm - z_i \partialf{\znorm}{z_j}}{\znorm^2} \\
      &= e^{\zdiv{i}} \cdot \left[ \frac{1}{\znorm} \cdot \partialf{z_i}{z_j} - \frac{z_i}{\znorm^2} \cdot \partialf{}{z_j} \left(\ksum z_k^2\right)^{1/2} \right] \,, \\
      \label{eq:lc_grad_single}
    \end{aligned}
  \end{equation}
\end{small}%
if $i=j$,
\begin{small}
  \begin{equation}
    \begin{aligned}
      \partialf{}{z_j} \left( e^{\zdiv{j}} \right) = e^{\zdiv{j}} \cdot \left( \frac{1}{\znorm} - \frac{z_j^2}{\znorm^3} \right) \,, \\
      \label{eq:lc_grad_single_eq}
    \end{aligned}
  \end{equation}
\end{small}%
otherwise, if $i\neq j$,
\begin{small}
  \begin{equation}
    \begin{aligned}
      \partialf{}{z_j} \left( e^{\zdiv{i}} \right) = e^{\zdiv{i}} \cdot \left( - \frac{z_i z_j}{\znorm^3} \right) \,.
      \label{eq:lc_grad_single_neq}
    \end{aligned}
  \end{equation}
\end{small}%
On the other hand, the gradient of the latter term can be deduced as:
\begin{small}
  \begin{equation}
    \begin{aligned}
      &\partialf{}{z_j} \left( \ksum e^{\zdiv{k}} \right) = \sum_{k\neq j} \partialf{}{z_j} e^{\zdiv{k}} + \partialf{}{z_j} e^{\zdiv{j}} \\
      &= \sum_{k\neq j} \left( e^{\zdiv{k}} \cdot \left( - \frac{z_k z_j}{\znorm^3} \right) \right) + e^{\zdiv{j}} \cdot \left( \frac{1}{\znorm} - \frac{z_j^2}{\znorm^3} \right) \\
      &= \frac{1}{\znorm} \cdot e^{\zdiv{j}} - \frac{z_j}{\znorm^3} \ksum z_k \cdot e^{\zdiv{k}} \,.
      \label{eq:lc_grad_sum}
    \end{aligned}
  \end{equation}
\end{small}%
Equipped with the results of Eq.~\eqref{eq:lc_grad_single_eq}--Eq.~\eqref{eq:lc_grad_sum}, we can now continue to calculate the derivations in Eq.~\eqref{eq:lc_grad_supp}. Specifically, if $i=j$,
\begin{small}
  \begin{equation}
    \begin{aligned}
      \partialf{p_i}{z_j} &= \left(\ksum e^{\zdiv{k}}\right)^{-2} \cdot \left[ e^{\zdiv{j}} \cdot \left( \frac{1}{\znorm} - \frac{z_j^2}{\znorm^3} \right) \ksum e^{\zdiv{k}} \right.\\
      &\left.\quad- e^{\zdiv{j}} \left( \frac{1}{\znorm} \cdot e^{\zdiv{j}} - \frac{z_j}{\znorm^3} \ksum z_k \cdot e^{\zdiv{k}} \right) \right] \\
      &= \left(\ksum e^{\zdiv{k}}\right)^{-2} \cdot e^{\zdiv{j}} \left[ \frac{1}{\znorm} \cdot \left(\ksum e^{\zdiv{k}} - e^{\zdiv{j}}\right) \right. \\
      &\left.\quad- \frac{z_j}{\znorm^3} \left(z_j \ksum e^{\zdiv{k}} - \ksum z_k e^{\zdiv{k}}\right) \right] \\
      &= \p{j} \left[ \frac{1}{\znorm} \cdot \left( 1 - \p{j} \right) + \frac{z_j}{\znorm^3} \ksum z_k \p{k} - \frac{z_j^2}{\znorm^3} \right] \,,
      \label{eq:lc_pred_grad_eq}
    \end{aligned}
  \end{equation}
\end{small}%
otherwise, if $i\neq j$,
\begin{small}
  \begin{equation}
    \begin{aligned}
      \partialf{p_i}{z_j} &= \left(\ksum e^{\zdiv{k}}\right)^{-2} \cdot \left[ e^{\zdiv{i}} \cdot \left( - \frac{z_i z_j}{\znorm^3} \right) \ksum e^{\zdiv{k}} \right.\\
      &\left.\quad- e^{\zdiv{i}} \left( \frac{1}{\znorm} \cdot e^{\zdiv{j}} - \frac{z_j}{\znorm^3} \ksum z_k \cdot e^{\zdiv{k}} \right) \right] \\
      &= \left(\ksum e^{\zdiv{k}}\right)^{-2} \cdot e^{\zdiv{i}} \left[ \frac{1}{\znorm} \cdot \left( - e^{\zdiv{j}}\right) \right.\\
      &\left.\quad- \frac{z_j}{\znorm^3} \left( z_i \ksum e^{\zdiv{k}} - \ksum z_k e^{\zdiv{k}}\right)\right] \\
      &= \p{i} \left[ \frac{1}{\znorm} (- \p{j}) + \frac{z_j}{\znorm^3}\ksum z_k \p{k} - \frac{z_i z_j}{\znorm^3} \right] \,.
      \label{eq:lc_pred_grad_neq}
    \end{aligned}
  \end{equation}
\end{small}%

Eq.~\eqref{eq:lc_pred_grad_eq} and Eq.~\eqref{eq:lc_pred_grad_neq} can be rearranged into:
\begin{small}
  \begin{equation}
    \partialf{\p{i}}{z_j} = \left\{
    \begin{aligned}
      & \p{j} \left[\frac{1}{\znorm}(1-\p{j}) + \frac{z_j}{\znorm^3} \ksum z_k \p{k} - \frac{{z_j}^2}{\znorm^3}\right], i = j \\
      & \p{i} \left[\frac{1}{\znorm}(-\p{j}) + \frac{z_j}{\znorm^3} \ksum z_k \p{k} - \frac{z_i z_j}{\znorm^3}\right], i\neq j \,.
    \end{aligned}
    \right. 
  \end{equation}
\end{small}%
Therefore, the gradient in Eq.~\eqref{eq:lc_grad_supp} can be calculated as:
\begin{small}
  \begin{equation}
    \begin{aligned}
        \partialf{\loss}{z_j} &= - y_j \cdot \frac{1}{\p{j}} \cdot \p{j} \left[\frac{1}{\znorm}(1-\p{j}) + \frac{z_j}{\znorm^3} \ksum z_k \p{k} - \frac{{z_j}^2}{\znorm^3}\right] \\
        &\quad- \sum_{k\neq j} y_k \cdot \frac{1}{\p{k}} \cdot \p{k} \left[\frac{1}{\znorm}(-\p{j}) + \frac{z_j}{\znorm^3} \ksum z_k \p{k} - \frac{z_k z_j}{\znorm^3}\right] \\
        &= - y_j \left[\frac{1}{\znorm}(1-\p{j}) + \frac{z_j}{\znorm^3} \ksum z_k \p{k} - \frac{{z_j}^2}{\znorm^3}\right] \\
        &\quad- \sum_{k\neq j} y_k \left[\frac{1}{\znorm}(-\p{j}) + \frac{z_j}{\znorm^3} \ksum z_k \p{k} - \frac{z_k z_j}{\znorm^3}\right] \\
        &= -\frac{1}{\znorm} y_j + \frac{1}{\znorm} \p{j} - \ksum \frac{z_j z_k}{\znorm^3} \p{k} + \ksum y_k \frac{z_j z_k}{\znorm^3} \\
        &= \frac{1}{\znorm} \left( (\p{j} - y_j) - \ksum \frac{z_j z_k}{\znorm^2} (\p{k} - y_k) \right) \,.
    \end{aligned}
  \end{equation}
\end{small}

\section{Additional Results}
\label{sec:additional_results}

\begin{table*}[!htbp]
  \centering
  \resizebox{\textwidth}{!}{
    \def\arraystretch{1.1}
    \begin{tabular}{ l | c c c c c c c c c c c c c c c c c c c | c }
      \Xhline{1.2pt}
      Method & \rotatebox{0}{road} & \rotatebox{0}{side.} & \rotatebox{0}{buil.} & \rotatebox{0}{wall} & \rotatebox{0}{fence} & \rotatebox{0}{pole} & \rotatebox{0}{light} & \rotatebox{0}{sign} & \rotatebox{0}{veg.} & \rotatebox{0}{terr.} & \rotatebox{0}{sky} & \rotatebox{0}{pers.} & \rotatebox{0}{rider} & \rotatebox{0}{car}& \rotatebox{0}{truck} & \rotatebox{0}{bus} & \rotatebox{0}{train} & \rotatebox{0}{mbike} & \rotatebox{0}{bike} & mIoU \\
      \hline
      \hline
      SegFormer & 68.4 & 25.7 & 64.0 & 24.1 & 18.6 & 44.8 & 54.5 & 44.5 & 73.2 & 32.4 & 75.6 & 45.9 & 17.2 & 74.6 & 38.5 & 37.6 & 41.0 & 24.2 & 19.7 & 43.4 \\
      \gr \bf\method & 88.5 & 57.6 & 81.9 & 41.2 & 35.2 & 58.0 & 72.8 & 57.5 & 71.7 & 39.3 & 82.1 & 62.2 & 36.2 & 87.1 & 82.6 & 86.6 & 84.1 & 41.6 & 44.9 & 63.7 \\
      \Xhline{1.2pt}
    \end{tabular}
  }

  \caption{Comparison results on \textbf{Cityscapes $\rightarrow$ ACDC semantic segmentation task}. IoU score of each class and the mIoU score are reported on {\bf ACDC validation set}. 
  } \label{table:acdc_validation_class}
\end{table*}

\begin{table*}[!htbp]
  \centering
  \resizebox{\textwidth}{!}{
    \def\arraystretch{1.1}
    \begin{tabular}{ l | c | c c c c c c c c c c c c c c c c c c c | c }
        \toprule[1.2pt]
        Method & Dataset & \rotatebox{0}{road} & \rotatebox{0}{side.} & \rotatebox{0}{buil.} & \rotatebox{0}{wall} & \rotatebox{0}{fence} & \rotatebox{0}{pole} & \rotatebox{0}{light} & \rotatebox{0}{sign} & \rotatebox{0}{veg.} & \rotatebox{0}{terr.} & \rotatebox{0}{sky} & \rotatebox{0}{pers.} & \rotatebox{0}{rider} & \rotatebox{0}{car}& \rotatebox{0}{truck} & \rotatebox{0}{bus} & \rotatebox{0}{train} & \rotatebox{0}{mbike} & \rotatebox{0}{bike} & mIoU \\
        \midrule
        \multirow{2.5}{*}{\bf\method} & IDD & 91.1 & 44.0 & 69.0 & 51.4 & 20.9 & 35.9 & 33.5 & 61.3 & 87.3 & 25.1 & 88.3 & 57.6 & 64.9 & 71.4 & 63.5 & 51.3 & 0.0 & 66.0 & 22.2 & 52.9 \\
        \cmidrule(lr){2-22}
        & Mapillary & 67.3 & 36.3 & 84.2 & 36.4 & 42.4 & 36.1 & 54.9 & 68.2 & 81.1 & 49.6 & 73.4 & 66.8 & 50.4 & 86.4 & 55.1 & 60.4 & 38.9 & 53.9 & 55.9 & 57.8 \\
        \bottomrule[1.2pt]
    \end{tabular}
  }

  \caption{Comparison results on \textbf{Cityscapes $\rightarrow$ IDD + Mapillary semantic segmentation task (19 classes)}. IoU score of each class and the mIoU score are reported on {\bf IDD validation set and Mapillary validation set, respectively}. 
  } \label{table:multi_validation_class}
\end{table*}

\begin{table*}[!htbp]
  \centering
  \def\arraystretch{1.1}
    \begin{tabular}{ l | c | c c c c c c c | c }
        \toprule[1.2pt]
        Method & Dataset & \rotatebox{0}{flat} & \rotatebox{0}{constr.} & \rotatebox{0}{object} & \rotatebox{0}{nature} & \rotatebox{0}{sky} & \rotatebox{0}{human} & \rotatebox{0}{vehicle} & mIoU \\
        \midrule
        \multirow{2.5}{*}{\bf\method} & IDD & 93.0 & 58.2 & 31.2 & 89.8 & 91.3 & 70.8 & 83.1 & 73.9 \\
        \cmidrule(lr){2-10}
        & Mapillary & 71.4 & 78.4 & 46.5 & 78.3 & 70.0 & 70.6 & 86.6 & 71.7\\
        \bottomrule[1.2pt]
    \end{tabular}
  \caption{Comparison results on \textbf{Cityscapes $\rightarrow$ IDD + Mapillary semantic segmentation task (7 super classes)}. IoU score of each class and the mIoU score are reported on {\bf IDD validation set and Mapillary validation set, respectively}. 
  } \label{table:multi_validation_class7}
\end{table*}

\subsection{More Quantitative results}
\label{sec:more_quantitative}

\paragraph{Detailed Results on ACDC Validation Set.}
For the Cityscapes $\rightarrow$ ACDC semantic segmentation task, we list the evaluation result on the ACDC validation set in Table~\ref{table:acdc_validation_class} to give a full picture of the comparison between our \method and SegFormer. Both results on test set and validation set consistently demonstrate the capability of our method.

\paragraph{Detailed Results on IDD + Mapillary Validation Set.}
The 7-class and 19-class results for the Cityscapes $\rightarrow$ IDD + Mapillary semantic segmentation task are shown in Table~\ref{table:multi_validation_class7} and Table~\ref{table:multi_validation_class}, respectively.

\begin{table}[!htbp]
  \centering
  \resizebox{0.35\textwidth}{!}{
      \begin{tabular}{ c | c c c c c }
          \toprule[1.2pt]
          $\alpha$ & 0.9 & 0.99 & 0.995 & 0.999 & 0.9995 \\
          \midrule
          mIoU & 62.3 & 63.2 & 63.4 & \bf 63.7 & 63.3 \\
          \bottomrule[1.2pt]
      \end{tabular}
  }
  \caption{Effect of ema update ratio $\alpha$.}
  \label{table:alpha}
\end{table}

\begin{table}[!htbp]
  \centering
  \resizebox{0.35\textwidth}{!}{
      \begin{tabular}{ c | c c c c c c }
          \toprule[1.2pt]
          $\delta$ & 0.7 & 0.8 & 0.85 & 0.9 & 0.95 \\
          \midrule
          mIoU & 63.3 & 62.7 & 63.2 & \bf 63.7 & 62.9 \\
          \bottomrule[1.2pt]
      \end{tabular}
  }
  \caption{Effect of pseudo threshold $\delta$.}
  \label{table:pseudo}
\end{table}

\begin{table}[!htbp]
  \centering
  \resizebox{0.35\textwidth}{!}{
      \begin{tabular}{ c | c c c c c c c }
          \toprule[1.2pt]
          $\gamma$ & 3.0 & 3.5 & 4.0 & 4.5 & 5.0 \\
          \midrule
          mIoU & 59.2 & 61.4 & \bf 63.7 & 63.8 & 62.8 \\
          \bottomrule[1.2pt]
      \end{tabular}
  }
  \caption{Effect of scaling factor $\gamma$.}
  \label{table:scale}
\end{table}

\subsection{Results on Hyperparameters}
\label{sec:parameters}

In this section, we explore the optimal values for the hyperparameters, and remark on the sensitivity of \method to them. All experiments are evaluated on ACDC validation set.

\paragraph{Delve into momentum-update ratio $\alpha$.}
Table \ref{table:alpha} shows the effect of different values of $\alpha$. It can be observed that the teacher model brings about a massive gain of over +5.0\% mIoU. The performance is relatively stable, and we fix $\alpha$ to 0.999 for a suitable regularization.

\paragraph{Delve into pseudo threshold $\delta$.}
We experiment on a variety of pseudo thresholds $\delta$, and list the results in Table \ref{table:pseudo}. The findings show that performance is relatively insensitive to $\delta$, and we decide to set $\delta$ to 0.9 for pursuit of a better model.

\paragraph{Delve into scaling factor $\gamma$.} To discover the optimal value for the global scaling factor for transmission map modulation, we run experiments with different values for $\gamma$. As is shown in Table~\ref{table:scale}, the best performance is gained around 4.0, and we fix $\gamma$ to this value.

\subsection{More Qualitative Results}
\label{sec:more_qualitative}

\paragraph{VBM versus Domain Transfer.} Although the adverse conditions could be caused by multiple factors, such as specific location, season, illumation, based on our observation, one of the primary causes is the visual appearance differences in the input space. In the literature, some works usually solve this problem via domain transfer strategies such as generative models or pixel-to-pixel translation models~\cite{hoffman2018cycada,li2019bidirectional} while another line of research engages in adopting Fourier transformation~\cite{yang2020fda}. These methods have proven that transferring image style of one domain to another domain can diminish the domain difference. Therefore, one pomissing way to improve the visiblity of adverse-condition images (target domain) is to transfer the style of clear-condition images (source domain).

\paragraph{More Segmentation Results.} 

\begin{figure}[H]
  \centering
  \includegraphics[width=0.47\textwidth]{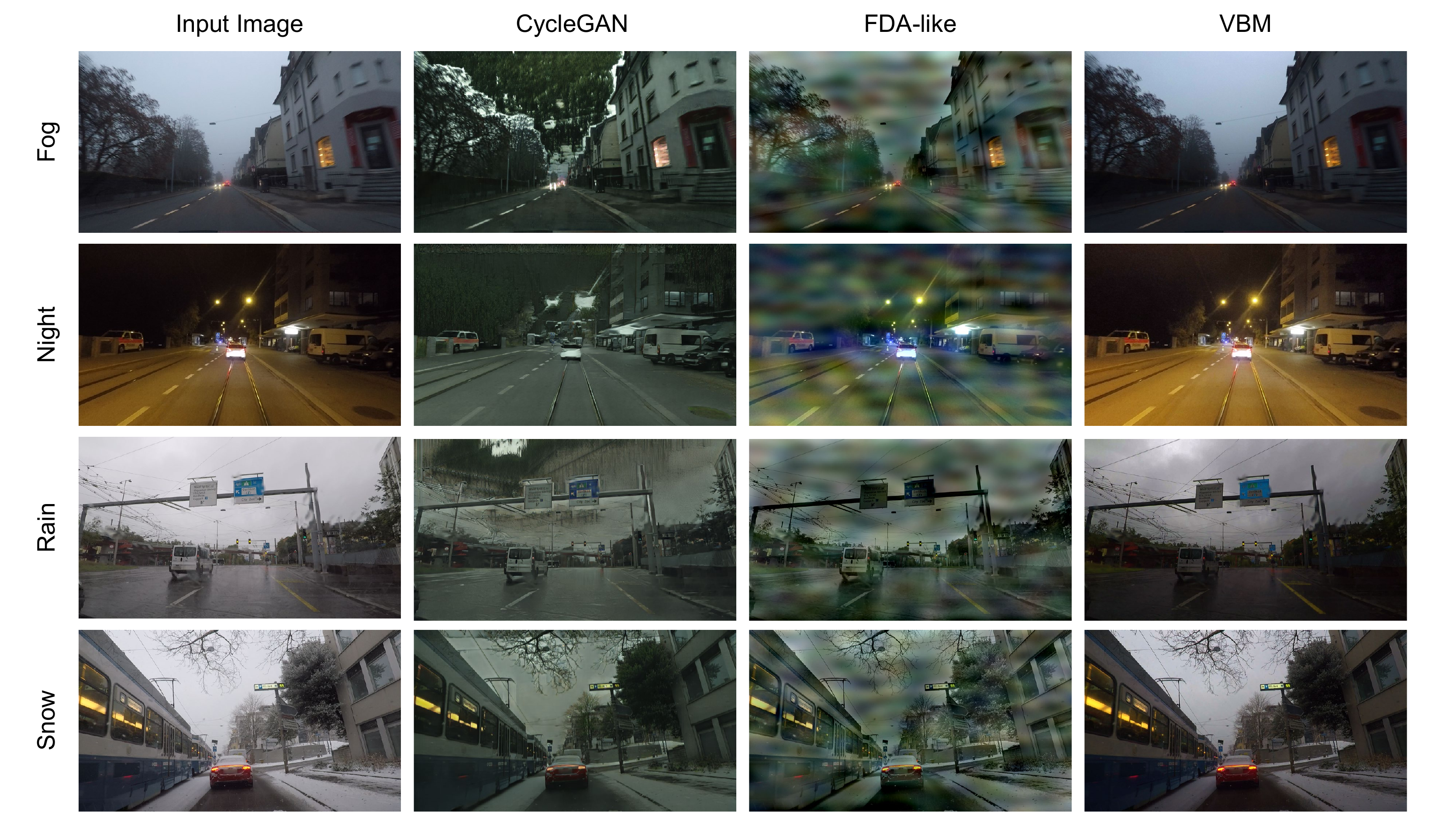}
  \caption[p]{Qualitative analysis on poor visibility image enhancement. From left to right: Input images, CycleGAN, FDA-like, and our VBM are shown one by one.}
  \label{fig:vbm_compare}
\end{figure}

Here, in Fig.~\ref{fig:vbm_compare}, we visualize some representative translated (boosted) images produced with CycleGAN, FDA-like, and our VBM, on the task of adapting from Cityscapes to ACDC. CycleGAN can generate the clear-condition images according to a randomly selected source domain reference image, while it requires to carry out a computationally expensive training process and would lost much more details. For example, the overall saturation is completely biased towards the source domain and the generated sky is also chaotic. As for FDA-like method, the concatenation of frequencies usually introduces significant noises during training, which largely limits its final performance. Our proposed VBM, in contrast, significantly increases the saturation of the image, thus improving the visibility of the image to some extent.

\begin{figure*}[!htbp]
  \centering
  \includegraphics[width=0.87\textwidth]{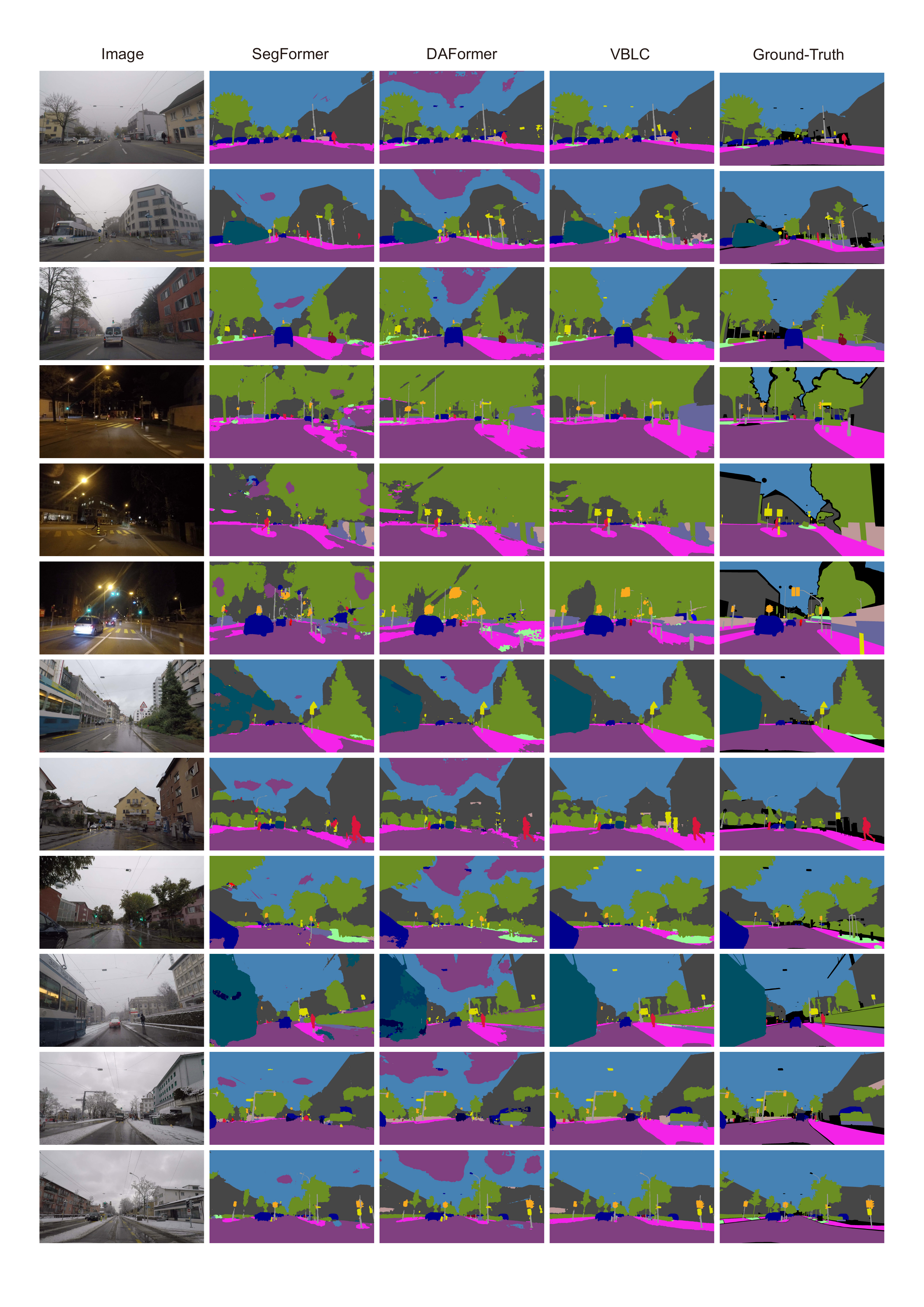}
  \caption{Qualitative analysis of segmentation results on ACDC validation set. From left to right: Target adverse-condition images, results predicted by SegFormer, by DAFormer, and by our \method, as well as ground-truth labels are shown one by one.}
  \label{fig:seg_map_supp}
\end{figure*}

We display more qualitative segmentation results on ACDC validation set in Fig.~\ref{fig:seg_map_supp}. It could be observed that our method generally produces finer predictions than SegFormer-based counterparts, with a clear visual improvement.

\begin{figure*}[!htbp]
  \centering
  \includegraphics[width=\textwidth]{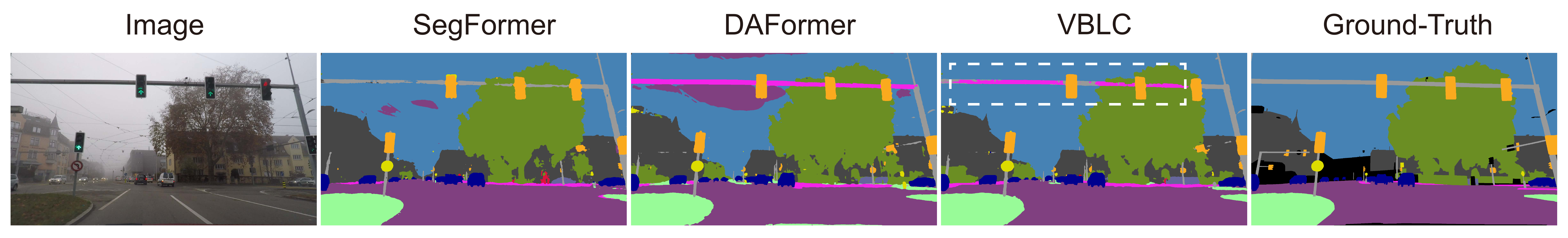}
  \caption{Typical error cases on Cityscapes $\rightarrow$ ACDC: Misclassification of horizontal poles.}
  \label{fig:failure_case_1}
\end{figure*}

\begin{figure*}[!htbp]
  \centering
  \includegraphics[width=\textwidth]{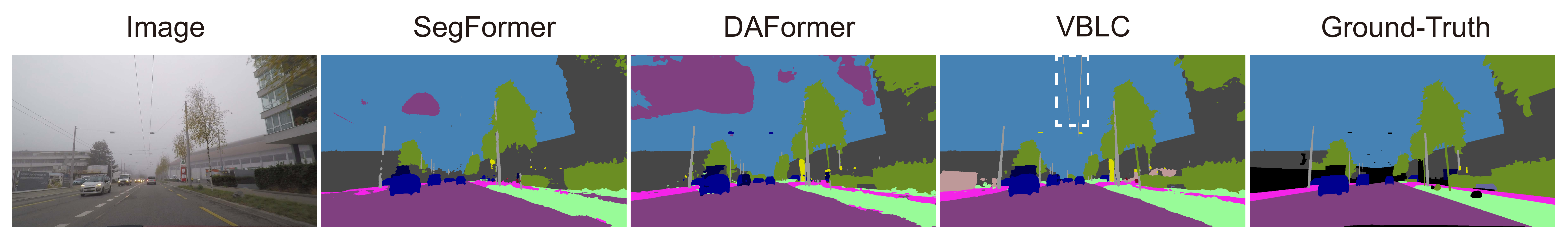}
  \caption{Typical error cases on Cityscapes $\rightarrow$ ACDC: Misclassification of overhead wires.}
  \label{fig:failure_case_2}
\end{figure*}

\begin{figure*}[!htbp]
  \centering
  \includegraphics[width=\textwidth]{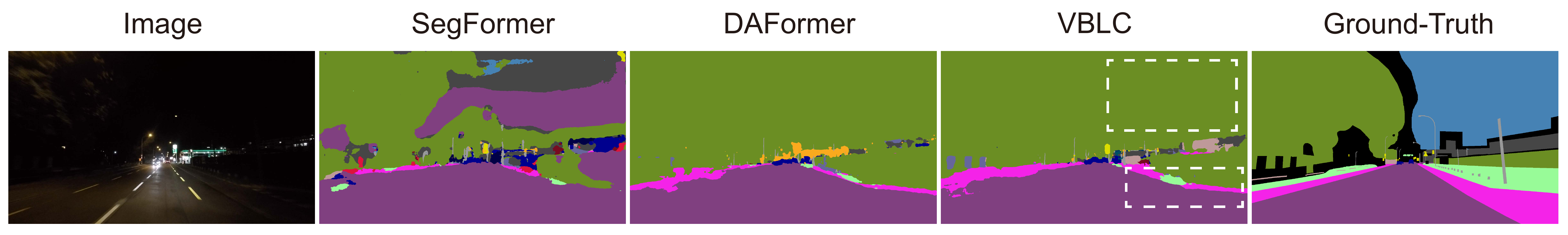}
  \caption{Typical error cases on Cityscapes $\rightarrow$ ACDC: Misclassification of sky and sidewalks due to darkness.}
  \label{fig:failure_case_3}
\end{figure*}

\begin{figure*}[!htbp]
  \centering
  \includegraphics[width=\textwidth]{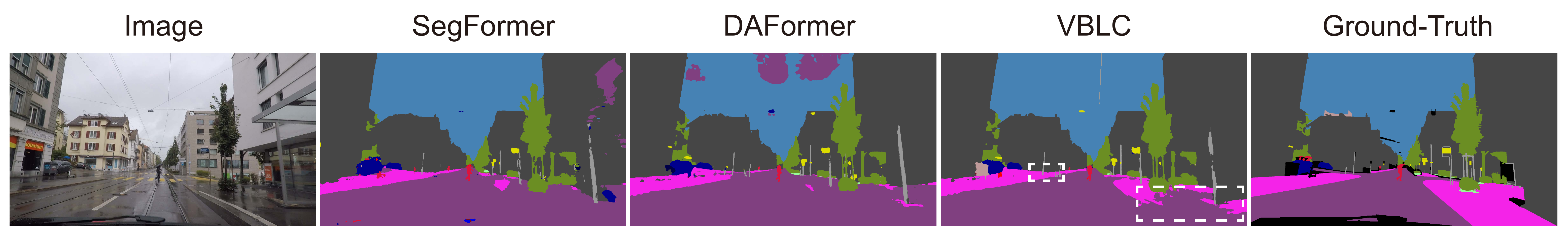}
  \caption{Typical error cases on Cityscapes $\rightarrow$ ACDC: Misclassification of sidewalks due to surface gathered water.}
  \label{fig:failure_case_4}
\end{figure*}

\begin{figure*}[!htbp]
  \centering
  \includegraphics[width=\textwidth]{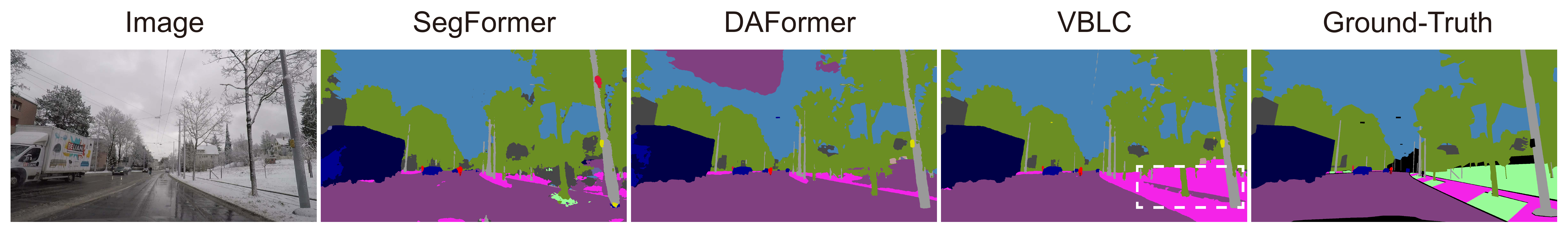}
  \caption{Typical error cases on Cityscapes $\rightarrow$ ACDC: Misclassification of terrain due to snow cover.}
  \label{fig:failure_case_5}
\end{figure*}

\subsection{Failure Cases}
\label{sec:failure}
Despite the improved robustness achieved by our \method, there still exist a few failure cases. Fig.~\ref{fig:failure_case_1}--\ref{fig:failure_case_5} list some typical failure cases we encounter. These cases include the wrong prediction of horizontal poles into the sidewalk (Fig.~\ref{fig:failure_case_1}), the discovery of overhead wires after visibility boost (Fig.~\ref{fig:failure_case_2}), incapability of predicting \textit{sky} and other classes in the extremely dark scene (Fig.~\ref{fig:failure_case_3}), confusion of \textit{sidewalk} and \textit{road} due to reflection by surface gathered water (Fig.~\ref{fig:failure_case_4}), and confusion of \textit{sidewalk} and \textit{terrain} caused by deep snow cover (Fig.~\ref{fig:failure_case_5}). Note that these failure cases are not exclusive to \method, but we would like to underline on them to motivate future work.

\end{document}